%% file: acl_latex.tex
\pdfoutput=1

\documentclass[11pt]{article}

\usepackage{acl}

\usepackage{times}
\usepackage{latexsym}

\usepackage[T1]{fontenc}

\usepackage[utf8]{inputenc}

\usepackage{microtype}

\usepackage{amsmath}
\usepackage{amssymb}
\usepackage{bm}
\usepackage{float}
\usepackage{multirow}
\usepackage{booktabs}
\usepackage{graphicx}
\usepackage[boxed]{algorithm2e}
\SetAlCapSty{}

\usepackage{array}
\usepackage{enumitem}
\usepackage{xcolor}
\usepackage{hhline}

\newcommand{\tabitem}{~~\llap{\textbullet}~~}

\title{Multilingual Detection of Personal Employment Status on Twitter}

\author{Manuel Tonneau$^{1,2,3}$, Dhaval Adjodah$^{1,2,4}$, João Palotti$^{5}$, \\ {\bf Nir Grinberg}$^{6}$ \and {\bf Samuel Fraiberger}$^{1,2,4}$\\
$^{1}$The World Bank\hspace{0.5cm} $^{2}$New York University\hspace{0.5cm} $^{3}$Centre Marc Bloch\\ $^{4}$Massachusetts Institute of Technology\hspace{0.5cm}
$^{5}$Qatar Computing Research Institute\hspace{0.5cm} \\ $^{6}$Ben-Gurion University of the Negev \\
}

\begin{document}
\maketitle
\begin{abstract}
Detecting disclosures of individuals' employment status on social media can provide valuable information to match job seekers with suitable vacancies, offer social protection, or measure labor market flows. However, identifying such personal disclosures is a challenging task due to their rarity in a sea of social media content and the variety of linguistic forms used to describe them. Here, we examine three Active Learning (AL) strategies in real-world settings of extreme class imbalance, and identify five types of disclosures about individuals' employment status (e.g. job loss) in three languages using BERT-based classification models. Our findings show that, even under extreme imbalance settings, a small number of AL iterations is sufficient to obtain large and significant gains in precision, recall, and diversity of results compared to a supervised baseline with the same number of labels. We also find that no AL strategy consistently outperforms the rest. Qualitative analysis suggests that AL helps focus the attention mechanism of BERT on core terms and adjust the boundaries of semantic expansion, highlighting the importance of interpretable models to provide greater control and visibility into this dynamic learning process.


\end{abstract}

\section{Introduction}
\label{intro}

Up-to-date information on individuals' employment status is of tremendous value for a wide range of economic decisions, from firms filling job vacancies to governments designing social protection systems. At the aggregate level, estimates of labor market conditions are traditionally based on nationally representative surveys that are costly to produce, especially in low- and middle-income countries~\cite{devarajan2013africa,jerven2013poor}. As social media becomes more ubiquitous all over the world, more individuals can now share their employment status with peers and unlock the social capital of their networks. This, in turn, can provide a new lens to examine the labor market and devise policy, especially in countries where traditional measures are lagging or unreliable. 

\begin{figure}[t]
    \raggedleft
    \includegraphics[width=0.45\textwidth]{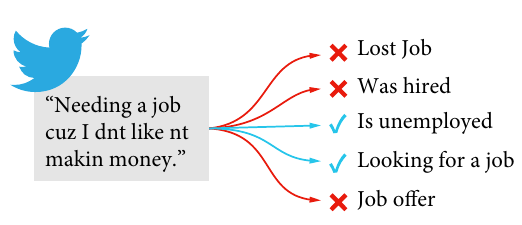} 
    \caption{\footnotesize{An example of a tweet suggestive of its author currently being unemployed and actively looking for a job.}}
    \label{fig:labeling_task_description}
\end{figure} 

A key challenge in using social media to identify personal disclosures of employment status is that such statements are extremely rare in an abundance of social media content -- roughly one in every 10,000 posts -- which renders random sampling ineffective and prohibitively costly for the development of a large labeled dataset. On the other hand, simple keyword-based approaches run the risk of providing seemingly high-accuracy classifiers while substantially missing linguistic variety used to describe events such as losing a job, looking for a job, or starting a new position (see Figure~\ref{fig:labeling_task_description} for example).
In the absence of a high-quality, comprehensive, and diverse ground-truth about personal employment disclosures, it is difficult to develop classification models that accurately capture the flows in and out of the labor market in any country, let alone robustly estimating it across multiple countries.
Furthermore, state-of-the-art deep neural models provide little visibility into or control over the linguistic patterns captured by the model, which hampers the ability of researchers and practitioners to determine whether the model has truly learned new linguistic forms and sufficiently converged. 

Active Learning (AL) is designed for settings where there is an abundance of unlabeled examples and limited labeling resources~\cite{cohn1994improving}. It aims to focus the learning process on the most informative samples and maximize model performance for a given labeling budget. In recent years, AL proved successful in several settings, including  policy-relevant tasks involving social media data~\cite{pohl2018batch,palakodety2020voice}.

The success of pre-trained language models such as BERT~\cite{devlin2019bert} in a variety of language understanding tasks has sparked interest in using AL with these models for imbalanced text classification. Yet, most research in this field has focused on artificially-generated rarity in data or imbalance that is not as extreme as the present setting \cite{ein-dor-etal-2020-active, schroder2021uncertainty}. Therefore, there is no evidence of the efficiency of AL using BERT-based models for sequence classification in real-world settings with extreme imbalance. It is unclear whether some AL strategies will perform significantly better than others in these settings, how quickly the different strategies will reach convergence (if at all),  and how the different strategies will explore the linguistic space.


In this work, we leverage BERT-based models~\cite{devlin2019bert} in three different AL paradigms to identify tweets that disclose an individual's employment status or change thereof. We train classifiers in English, Spanish, and Portuguese to determine whether the author of a tweet recently lost her job, was recently hired, is currently unemployed, posting to find a job, or posting a job offer. We use two standard AL strategies, Uncertainty Sampling~\cite{lewis1994sequential} and Adaptive Retrieval~\cite{mussmann-etal-2020-importance}, and propose a novel strategy we name Exploit-Explore Retrieval that uses k-skip-n-grams (n-grams with $k$ skipped tokens) to explore the space and provide improved interpretability. We evaluate the models both quantitatively and qualitatively across languages and AL strategies, and compare them to a supervised learning baseline with the same number of labels. Therefore, our contributions are:

\begin{itemize}[leftmargin=*, topsep=0pt]
\setlength{\parskip}{0pt}
\setlength{\itemsep}{0pt plus 1pt}
    \item An evaluation of three AL strategies for BERT-based binary classification under extreme class imbalance using real-world data. 
    
    \item A novel AL strategy for sequence classification that performs on par with other strategies, but provides additional interpretability and control over the learning process.
    
    \item A qualitative analysis of the linguistic patterns captured by BERT across AL strategies.
    

    \item A large labeled dataset of tweets about unemployment and fine-tuned models in three languages to stimulate research in this area\footnote{Labeled datasets and models can be found at \url{https://github.com/manueltonneau/twitter-unemployment}}.
    
\end{itemize}

\section{Background and related work} \label{2_related_work}

\subsection{Identifying self-disclosures on Twitter}


Social media users disclose information that is valuable for public policy in a variety of areas ranging from health~\cite{achrekar2011predicting, mahata2018detecting, klein2018social} to emergency response to natural disasters~\cite{bruns2012tools, kryvasheyeu2016rapid} through migration flows~\cite{fiorio2017using, chi2020general, palotti2020monitoring}.
A key challenge in identifying self-disclosures on social media is the rare and varied nature of such content with a limited labeling budget. Prior work that studied self-disclosures on Twitter had either used pattern matching, which is prone to large classification errors~\cite{antenucci2014using,proserpio2016psychology}, or focused on curated datasets~\cite{li-etal-2014-major, preotiuc-pietro-etal-2015-analysis, sarker2018data, ghosh-chowdhury-etal-2019-speak}, which provide no guarantees about recall or coverage of the positive class. These issues are more severe in real-world settings of extreme imbalance, where random sampling is unlikely to retrieve any positives, let alone diverse. These challenges motivate the use of AL, as described next.

\subsection{Active Learning}
\label{2_3_active_learning}

AL has been used successfully in various settings to maximize classification performance for a given labeling budget (see~\citet{settles1995active} for a survey). 
With the emergence of pre-trained language models such as BERT \cite{devlin2019bert} and their success across a number of different language tasks, recent work has studied the combination of AL and BERT, either by using BERT to enhance traditional AL methods \cite{yuan2020alps} or by applying established AL methods to improve BERT's classification performance \cite{zhang2019ensemble,shelmanov2019active,liu2020ltp,griesshaber-etal-2020-fine,prabhu2021multi,schroder2021uncertainty}. 

In the specific case of binary classification with moderate class imbalance, \citet{ein-dor-etal-2020-active} show that AL with BERT significantly outperforms random sampling but that no single AL strategy stands out in terms of BERT-based classification performance, both for balanced and imbalanced settings. Yet, the authors only consider a relatively moderate class imbalance of 10-15\% positives, and does not cover extreme imbalance, which is common in many text classification tasks. Our current research examines a considerably more extreme imbalance of about 0.01\% positives, where traditional AL approaches can be ineffective~\cite{attenberg2010label}. Under this extreme imbalance, \citet{mussmann-etal-2020-importance} show the potential of AL for BERT to outperform random sampling for pairwise classification. To the best of our knowledge, this work is the first to compare the performance of AL methods for BERT-based sequence classification in real-world extreme imbalance settings.

\section{Experimental procedure} \label{3_experimental_procedure}

\subsection{Data collection}

Our dataset was collected from the Twitter API. It contains the timelines of the users with at least one tweet in the Twitter Decahose and with an inferred profile location in the United States, Brazil, and Mexico. In addition to the United States, we chose to focus on Brazil and Mexico as both of them are middle-income countries where Twitter's penetration rate is relatively high. For each country, we drew a random sample of 200 million tweets covering the period between January 2007 and December 2020 and excluding retweets. We then split it evenly in two mutually exclusive random samples $R_e$ and $R_s$. In the following sections, we use $R_e$ to evaluate each model's performance in a real-world setting and $R_s$ to sample new tweets to label.

Our labeling process sought to identify four non-exclusive, binary states that workers may experience during their career: losing a job (``Lost Job''), being unemployed (``Is Unemployed''), searching for a job (``Job Search''), and finding a job (``Is Hired''). We only considered first-person disclosures as positives. For the classes ``Lost Job'' and ``Is Hired'', we only considered such events that happened in the past month as positives as we want to determine the user's current employment status.  To complement the focus on workers, we also labeled tweets containing job offers ("Job Offer"). We used Amazon Mechanical Turk (MTurk) to label tweets according to these 5 classes (see Figure~\ref{fig:labeling_task_description} and Section~\ref{appendix_data_labeling} for details).


\subsection{Initialization sample}
\label{3_2_initialization_sample}

As previously stated, the extreme imbalance of our classification task of one positive example for every 10,000 tweets renders random sampling ineffective and prohibitively costly. In order to build high-performing classifiers at a reasonable cost, we selected a set of 4 to 7 seed keywords that are highly specific of the positives and frequent enough for each class and country. To do so, we defined a list of candidate seeds, drawing from \citet{antenucci2014using} for the US and asking native speakers in the case of Mexico and Brazil, and individually evaluated their specificity and frequency (see  Section~\ref{stratified_sampling} for additional details). We then randomly sampled 150 tweets containing each seed from $R_s$, allowing us to produce a stratified sample $L_0$ of 4,524 English tweets, 2703 Portuguese tweets, and 3729 Spanish tweets respectively (Alg.~\ref{fig:bigpicture}). We then labeled each tweet using Amazon Mechanical Turk (MTurk) allowing us to construct a language-specific stratified sample that is common to the 5 classes (see Section~\ref{appendix_dataset_description} for descriptive statistics of the stratified sample). 







\subsection{Models}
\label{finetuning_evaluation}

We trained five binary classifiers to predict each of the five aforementioned labeled classes. Preliminary analysis found that BERT-based models considerably and consistently outperformed keyword-based models, static embedding models, and the combination of these models. We benchmarked several BERT-based models and found that the following models gave the best performance on our task: \textbf{Conversational BERT} for English tweets ~\cite{burtsev2018deeppavlov}, \textbf{BERTimbau} for Brazilian Portuguese tweets~\cite{souza2020bertimbau} and \textbf{BETO} for Mexican Spanish tweets~\cite{CaneteCFP2020} (see Section~\ref{appendix_model_characteristics} for details on model selection).

We fine-tuned each BERT-based model on a 70:30 train-test split of the labeled tweets for 20 epochs (Alg.~\ref{fig:bigpicture}). Following \citet{dodge2020fine}, we repeated this process for 15 different random seeds and retained the best performing model in terms of area under the ROC curve (AUROC) on the test set at or after the first epoch (see Section~\ref{appendix_finetuning_evaluation} for details).

\subsection{Model evaluation}
\label{evaluation_metrics}

While the standard classification performance measure in an imbalanced setting is the F1 score with a fixed classification threshold (e.g. 0.5), it is not applicable in our case for two reasons. First, we care about the performance on a large random set of tweets and the only labeled set we could compute the F1 metric from is the stratified test set which is not representative of the extremely imbalanced random sample $R_e$. Second, the fact that neural networks are poorly calibrated \cite{guo2017calibration} makes the choice of a predefined classification threshold somewhat arbitrary and most likely sub-optimal. 

We developed an alternative threshold-setting evaluation strategy. First, we computed the predicted score of each tweet in $R_e$ (Alg.~\ref{fig:bigpicture}), which is a random sample. Then, for each class, we labeled 200 tweets in $R_e$ along the score distribution (see section \ref{appendix_evaluation_sampling} for more details). 
We measured the performance of each classifier on $R_e$ by computing:

\begin{itemize}[leftmargin=*, topsep=0pt]
\setlength{\parskip}{0pt}
\setlength{\itemsep}{0pt plus 1pt}
    \item the \textbf{Average Precision} as common in information retrieval. 
    \item the \textbf{number of predicted positives}, defined as the average rank in the confidence score distribution when the share of positives reaches 0.5. 
    \item the \textbf{diversity}, defined as the average pairwise distance between true positives.
\end{itemize}

\noindent Details about the evaluation metrics can be found in Section \ref{appendix_evaluation_metrics}. 

\input{tables/algorithm_experimental_procedure}

\subsection{Active Learning strategies}
\label{description_al_methods}

Next, we used pool-based AL \cite{settles1995active} in batch mode, with each class-specific fine-tuned model as the classification model, in order to query new informative tweets in $R_s$. We compared three different AL strategies aiming to balance the goal of improving the precision of a classifier while expanding the number and the diversity of detected positives instances:

\begin{itemize}[leftmargin=*, topsep=0pt]
\setlength{\parskip}{0pt}
\setlength{\itemsep}{0pt plus 1pt}
\item \textbf{Uncertainty Sampling} consists in sampling instances that a model is most uncertain about. In a binary classification problem, the standard approach is to select examples with a predicted score close to 0.5 \cite{settles2009active}. In practice, this rule of thumb might not always lead to identify uncertain samples when imbalance is high \cite{mussmann-etal-2020-importance}, especially with neural network models known to be poorly calibrated \cite{guo2017calibration}. To overcome this issue, we contrast a naive approach which consists in querying the 100 instances whose \textit{uncalibrated} scores are the closest to 0.5, to an approach that uses \textit{calibrated} scores (see Section~\ref{appendix_calibration} for details).

\item \textbf{Adaptive Retrieval} aims to maximize the precision of a model by querying instances for which the model is most confident of their positivity \cite{mussmann-etal-2020-importance}. This approach is related to certainty sampling \cite{attenberg2010unified}. Here, we select the 100 tweets whose predicted score is the highest for each class.

\item Our novel strategy, \textbf{Exploit-Explore Retrieval} (see Section~\ref{appendix_algorithm} for details), aims to maximize precision (`exploitation') while improving recall by feeding new and diverse instances at each iteration (`exploration'):
\begin{itemize}[leftmargin=*]
    \setlength{\parskip}{0pt}
    \setlength{\itemsep}{0pt plus 1pt}
    \item \textbf{Exploitation}: Randomly query 50 new tweets from the top $10^4$ tweets with the highest predicted score in $R_s$.
    \item \textbf{Exploration}: Identify the 10 k-skip-n-grams with the highest frequency of occurrences in the top $10^4$ tweets, relative to their frequency in $R_s$. Then, randomly sample 50 new tweets containing each k-skip-n-gram (see Section \ref{appendix_algorithm} for formal definition of k-skip-n-grams and a discussion on the choice of threshold).
\end{itemize}

\end{itemize}


Additionally, we compared these AL strategies to a supervised \textbf{Stratified Sampling} baseline, that consists of the same initial motifs defined in Section~\ref{3_2_initialization_sample} and the same number of labels as available to all other AL strategies.
Overall, for each strategy, each iteration and each class, we labeled 100 new tweets in $R_s$. We then combined the 500 new labels across classes with the existing ones to fine-tune and evaluate a new BERT-based model for each class as described in Section~\ref{finetuning_evaluation}, which we then used to select tweets for labeling for the next iteration. We considered that an AL strategy had converged when there was no significant variation of average precision, number of predicted positives and diversity for at least two iterations (see Section \ref{convergence} for details).

\section{Results} \label{4_results}


\subsection{Initial sample}
\label{results_iter0}
At iteration 0, we fine-tuned a BERT-based classifier on a 70:30 train-test split of the initialization sample $L_0$ for each class and country. All the AUROC values on the test set are reported in Table \ref{table_auc_iter0}. 

We obtain very high AUROCs ranging from 0.944 to 0.993 across classes and countries. 
``Job Offer'' has the highest AUROCs with values ranging from 0.985 for English to 0.991 for Portuguese and 0.993 for Spanish. Upon closer examination of positives for this class, we find that the linguistic structure of tweets mentioning job offers is highly repetitive, a large share of these tweets containing sentences such as ``We're \#hiring! Click to apply:'' or naming job listing platforms (e.g: ``\#CareerArc''). By contrast, the most difficult class to predict is ``Lost Job'',  with an AUROC on the test set equal to 0.959 for English and 0.944 for Spanish. This class also has the highest imbalance, with approximately 6\% of positives in the stratified sample for these two languages. 
\setlength{\parskip}{0pt}
\setlength{\itemsep}{0pt plus 1pt}

Taken together, these results show that a fine-tuned BERT model can achieve very high classification performance on a stratified sample of tweets across classes and languages. However, these numbers cannot be extrapolated to directly infer the models' performance on random tweets, which we discuss in the next section.





\subsection{Active Learning across languages}
\label{country_comparison}

\begin{figure*}[t]
    \raggedleft
    \includegraphics[width=1\textwidth]{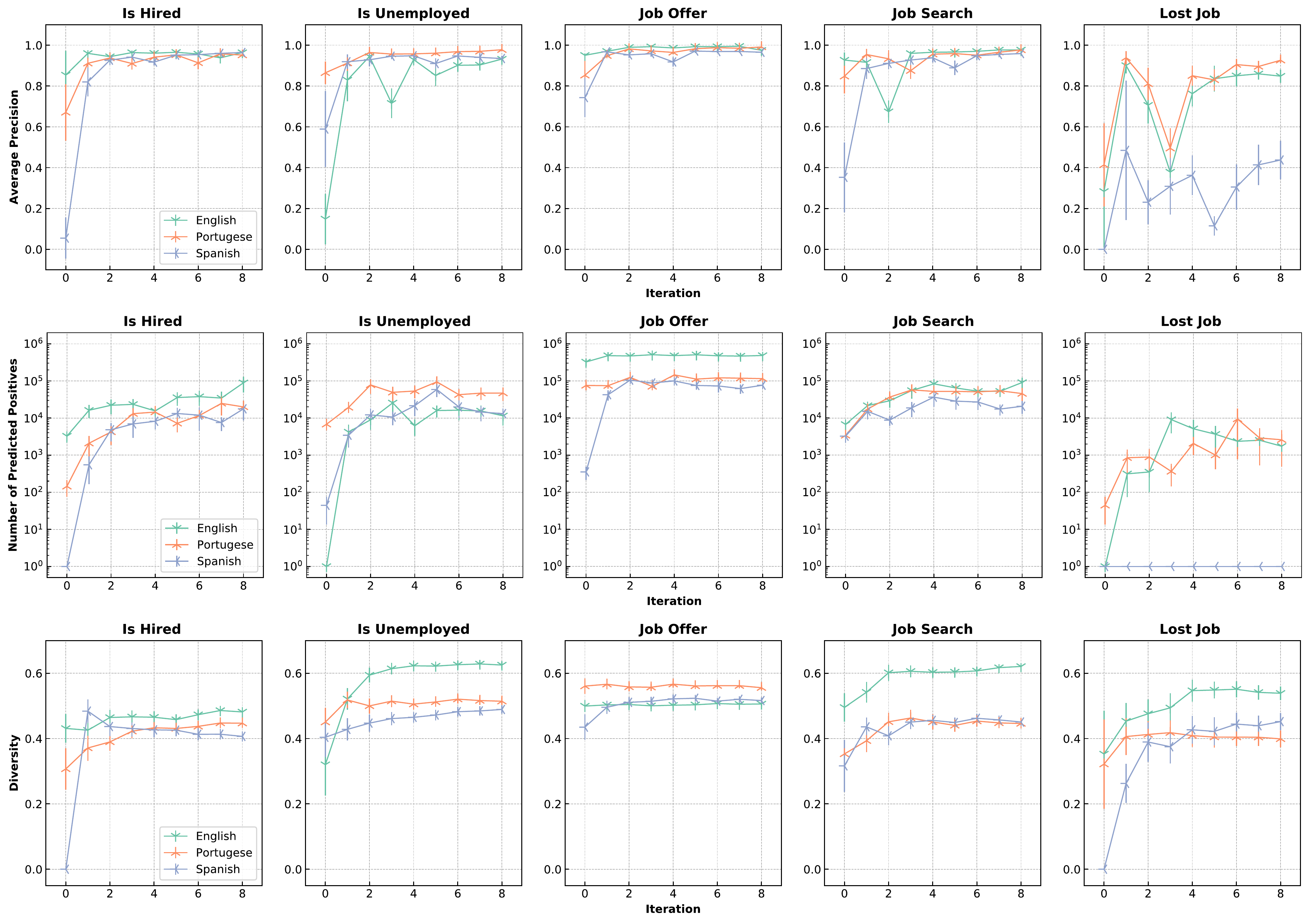} 
    \caption{\footnotesize{Average precision, number of predicted positives and diversity of true positives (in row) for each class (in column) for English (green), Portuguese (orange), and Spanish (purple). We report the standard error of the average precision and diversity estimates, and we report a lower and an upper bound for the number of predicted positives. Additional details on how the evaluation metrics are computed are reported in section \ref{appendix_evaluation_metrics}.}}
    \label{fig:masterplot_cross_country}
\end{figure*} 


Next, we compared the performance of our \textbf{exploit-explore retrieval} strategy on English, Spanish and Portuguese tweets. We used exploit-explore retrieval as it provides similar results to other strategies (Section~\ref{comparison_al_methods}), while allowing greater visibility into  selected motifs during the development process (Section~\ref{qualitative_analysis}). We ran 8 AL iterations for each language and report the results in Fig.~\ref{fig:masterplot_cross_country}, Fig.~\ref{fig:precision_country_comparison} and Table \ref{tab:avg_precision_cross_country}.

First, we observe substantial improvements in average precision (AP) across countries and classes with just one or two iterations.
These improvements are especially salient in cases where precision at iteration 0 is very low. For instance, for the English ``Is Unemployed'' class and the Spanish ``Is Hired'' class, average precision goes respectively from 0.14 and 0.07 to 0.83 and 0.8  from iteration 0 to iteration 1 (Fig.~\ref{fig:masterplot_cross_country} and Fig.~\ref{fig:precision_country_comparison}). A notable exception to this trend is the class ``Job Offer'', especially for English and Portuguese.
These performance differences can in part be explained by the varying quality of the initial seed list across classes. This is confirmed by the stratified sampling baseline performance discussed in \ref{comparison_al_methods}. In the case of ``Job Offer'', an additional explanation discussed earlier in Section~\ref{results_iter0} is the repetitive structure of job offers in tweets which makes this class easier to detect compared to others. 

Also, the class ``Lost Job'' has the worst performance in terms of AP across countries. One reason is that the data imbalance for this class is even higher than for other classes, as mentioned in Section~\ref{results_iter0}. Another explanation for the low precision is the ambiguity inherent to the recency constraint, namely that an individual must have lost her job at most one month prior to posting the tweet. 

Apart from the ``Job Offer'' class in English and Portuguese, AL consistently allows to quickly expand from iteration 0 levels with the number of predicted positives multiplied by a factor of up to $10^{4}$ (Fig.~\ref{fig:masterplot_cross_country}). Combined with high AP values, this result means that the classifiers manage to capture \emph{substantially} more positives compared to iteration 0. This high expansion is combined with increasing semantic diversity among true positive instances. 

The class ``Job Offer'' stands out with little expansion and diversity changes in the English and Portuguese cases. For Spanish, expansion and diversity changes are higher. One explanation is that the structure of Mexican job offers is less repetitive, with individual companies frequently posting job offers, as opposed to job aggregators in the case of the US and Brazil.

\setlength{\parskip}{0pt}
\setlength{\itemsep}{0pt plus 1pt}

Overall, apart from a few edge cases, we find that AL used with pre-trained language models is successful at significantly improving precision while expanding the number and the diversity of predicted positive instances in a small number of iterations across languages. Indeed, precision gains reach up to 90 percentage points from iteration 0 to the last iteration across languages and classes and the number of predicted positives is multiplied by a factor of up to $10^{4}$. Furthermore, on average, the model converges in only 5.6 iterations across classes for English and Portuguese, and in 4.4 iterations for Spanish (see Table \ref{tab:avg_precision_cross_country} for details).



\subsection{Comparing Active Learning strategies}
\label{comparison_al_methods}

\begin{figure*}[!ht]
    \centering
    \includegraphics[width=1\textwidth]{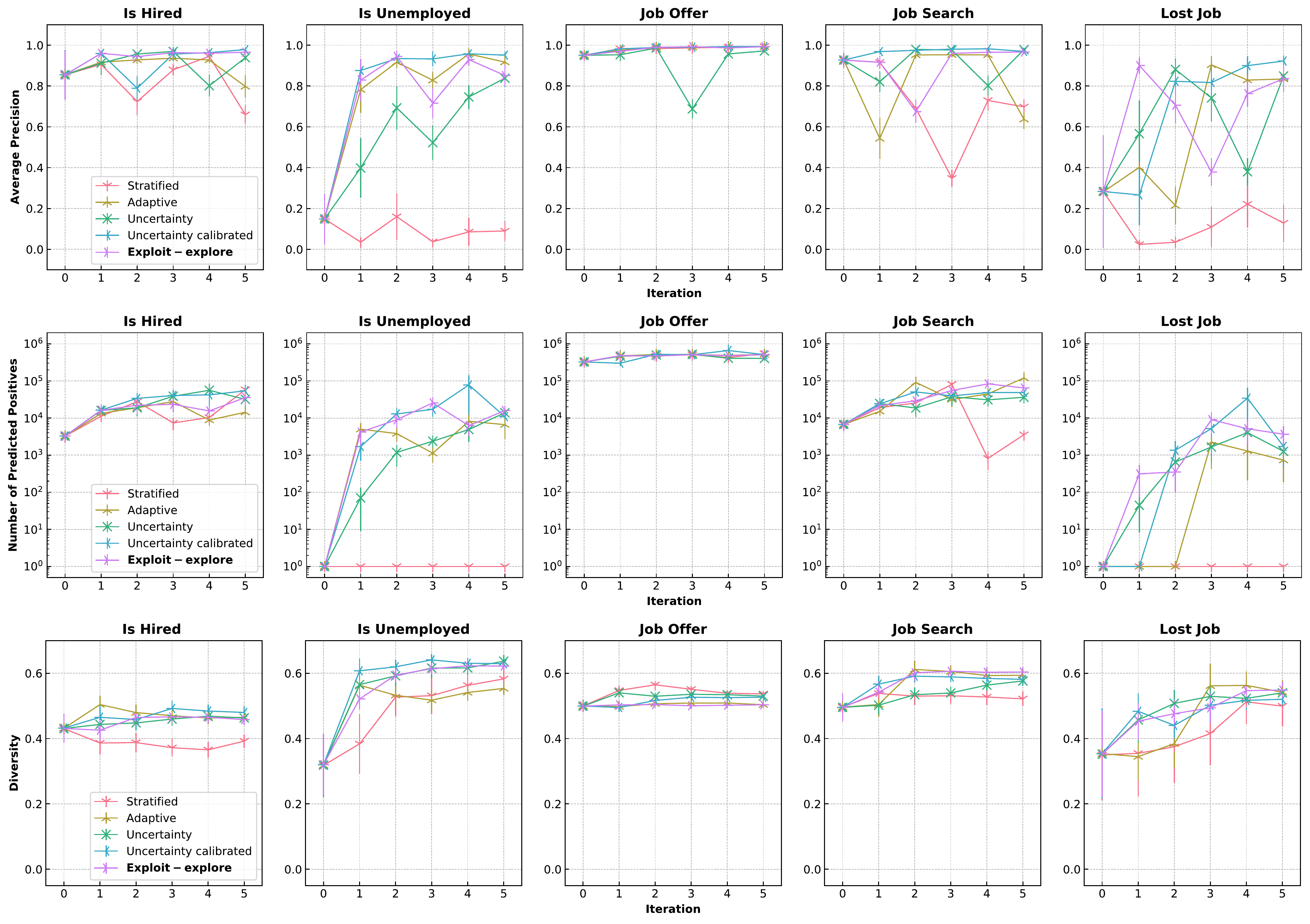} 
    \caption{\footnotesize{Average precision, number of predicted positives and diversity of true positives (in row) for each class (in column) across AL strategies. We report the standard error of the average precision and diversity estimates, and we report a lower and an upper bound for the number of predicted positives. Additional details on how the evaluation metrics are computed are reported in section \ref{appendix_evaluation_metrics}.}}
    \label{fig:masterplot_al_strategy}
\end{figure*} 


In this section, we evaluated on English tweets the stratified sampling baseline and the four AL strategies described in Section~\ref{description_al_methods}, namely exploit-explore retrieval, adaptive retrieval and uncertainty sampling with and without calibration. We ran five iterations for each strategy and reported the results on Figure~\ref{fig:masterplot_al_strategy} in this section as well as Table \ref{tab:avg_precision_cross_method} and Figure~\ref{fig:precision_al_method_comparison} in Section~\ref{appendix_additional_results_cross_country}.

We find that AL brings an order of magnitude more positives and does so while preserving or improving both the precision and the diversity of results. Apart from the ``Job Offer'' class discussed in Section \ref{country_comparison}, AL consistently outperforms the stratified sampling baseline. This is especially true for the classes ``Is Unemployed'' and ``Lost Job'' where the baseline performance stagnates at a low level, suggesting a poor seed choice, but also holds for classes ``Is Hired'' and ``Job Search'' with stronger baseline performance. We also find that no AL strategy consistently dominates the rest in terms of precision, number and diversity of positives. The gains in performance are similar across AL strategies, and are particularly high for the classes ``Lost Job'' and ``Is Unemployed'', which start with a low precision. The number of predicted positives and the diversity measures also follow similar trends across classes and iterations.

We also observe occasional ``drops'' in average precision of more than 25\% from one iteration to the next. Uncalibrated uncertainty sampling seems particularly susceptible to these drops, with at least one occurrence for each class. Upon examination of the tweets sampled for labeling by this strategy, the vast majority of tweets are negatives and when a few positives emerge, their number is not large enough to allow the model to generalize well. This variability slows down the convergence process of uncertainty sampling when it is not uncalibrated (table \ref{tab:avg_precision_cross_method}). In contrast, calibrated uncertainty sampling is less susceptible to these swings, emphasizing the importance of calibration for more ``stable'' convergence in settings of extreme imbalance. 

Taken together, our quantitative results show that the positive impact of AL on classification performance in an extremely imbalanced setting holds across AL strategies. Aside from a few occasional performance ``drops'', we find significant gains in precision, expansion and diversity across strategies. Yet, we find that no AL strategy consistently dominates the others across a range of prediction tasks for which the number and the linguistic complexity of positive instances vary widely. Next, we investigate the results qualitatively to gain deeper understanding of the learning process.



\subsection{Qualitative analysis}
\label{qualitative_analysis}

We qualitatively examined the tweets selected for labeling by each strategy to understand better what BERT-based models capture and reflect on the quantitative results. We focused on English tweets only and took a subsample of tweets at each iteration to better understand each strategy's performance. We excluded the ``Job Offer'' class from this analysis since the performance, in this case, is exceptionally high, even at iteration 0. 

Our analysis finds that many tweets queried by the various AL strategies capture a general ``tone'' that is present in tweets about unemployment, but that is not specific to one's employment status. For example, these include tweets of the form of ``I'm excited to ... in two days'' for the recently hired class, ``I've been in a shitty mood for ...'' for unemployment or ``I lost my ...'' for job loss. This type of false positives seems to wane down as the AL iterations progress, which suggests that a key to the success of AL is first to fine-tune the attention mechanism to focus on the core terms and not the accompanying text that is not specific to employment status. In the stratified sampling case, the focus on this unemployment ``tone'' remains uncorrected, explaining the poor performance for classes ``Lost Job'' and ``Is Unemployed'' and the performance drops for ``Is Hired'' and ``Job Search''. 

A second theme in tweets queried by AL involves the refinement of the initial motifs. Uncertainty sampling (calibrated and uncalibrated), adaptive retrieval, and the exploitation part of our exploit-explore retrieval method seem to query tweets that either directly contain a seed motif or a close variant thereof. For example, tweets for the class ``Lost Job'' may contain the seed motifs ``laid off'', ``lost my job'', and ``just got fired''. As mentioned in Section~\ref{country_comparison} to explain occasional drops in performance, many tweets labeled as negatives contain over-generalization of the semantic concept such as expanding to other types of losses (e.g. ``lost my phone''), other types of actions (e.g. ``got pissed off''), or simply miss the dependence on first-person pronouns (e.g. ``@user got fired''). Many of the positively labeled tweets contain more subtle linguistic variants that do not change the core concept such as ``I \emph{really} need a job'', ``I really need to get a job'', ``I need \emph{to find} a job'', or ``I need a \emph{freaken} job''. Adaptive retrieval chooses these subtle variants more heavily than other strategies with some iterations mostly populated with ``I need a job'' variants. Overall, these patterns are consistent with a view of the learning process, specifically the classification layer of the BERT model, as seeking to find the appropriate boundaries of the target concept. 

Finally, the exploration part of the exploit-explore retrieval makes the search for new forms of expression about unemployment more explicit and interpretable. For example, the patterns explored in the first few iterations of explore-exploit retrieval include ``I ... lost ... today'', ``quit .. my .. job'', ``I ... start my ... today'', and ``I'm ... in ... need''. A detailed presentation of the explored k-skip-n-grams for US tweets can be found in Table~\ref{tab:emerging_motifs_US} of Section~\ref{appendix_algorithm}. While this strategy suffers from issues that also affect other AL strategies, we find that the explore part of exploit-explore retrieval is more capable of finding new terms that were not part of the seed list (e.g., quit, career) and provides the researcher with greater insight into and control over the AL process.




\section{Discussion and conclusion}
\label{5_discuss}


This work developed and evaluated BERT-based models in three languages and used three different AL strategies to identify tweets related to an individual's employment status. Our results show that AL achieves large and significant improvements in precision, expansion, and diversity over stratified sampling with only a few iterations and across languages. In most cases, AL brings an order of magnitude more positives while preserving or improving both the precision and diversity of results. Despite using fundamentally different AL strategies, we observe that no strategy consistently outperforms the rest. Within the extreme imbalance setting, this is in line with -- and complements -- the findings of \citet{ein-dor-etal-2020-active}. 

Additionally, our qualitative analysis and exploration of the exploit-explore retrieval give further insights into the performance improvements provided by AL, finding that substantial amounts of queried tweets hone the model's focus on employment rather than surrounding context and expand the variety of motifs identified as positive. This puts exploit-explore retrieval as a valuable tool for researchers to obtain greater visibility into the AL process in extreme imbalance cases without compromising on performance.


While the present work demonstrates the potential of AL for BERT-based models under extreme imbalance, an important direction for future work would be to further optimize the AL process. One could for instance study the impact on performance of the stratified sample size or the AL batch size. To overcome the poor seed quality for some classes, other seed generation approaches could be tested, such as mining online unemployment forums using topic modeling techniques to discover different ways to talk about unemployment. In terms of model training and inference, the use of multi-task learning for further performance improvement could be studied due to the fact that classes of unemployment are not mutually exclusive. We hope that our experimental results as well as the resources we make available will help bridge these gaps in the literature.

\section*{Ethics statement} 
We acknowledge that there is some risk, like any other technology that makes inferences at the individual level, that the technology presented here will be used for harm. However, due to the public nature of the content and the fact that the potential harm already exists using basic keyword search, we believe that the marginal risk added by our classifier is minimal.

\section*{Acknowledgements}
 We thank participants of the Israeli Seminar on Computational Linguistics at Ben-Gurion University of the Negev as well as the anonymous reviewers for their valuable comments. We also thank Aleister Montfort, Varnitha Kurli Reddy and Boris Sobol for their excellent research assistance. This work was supported by the SDG Partnership Fund. 

\bibliography{anthology,custom}
\bibliographystyle{acl_natbib}

\clearpage

\appendix

\section{Experimental details}

\subsection{Stratified sampling}
\label{stratified_sampling}

\input{tables/table_motifs}

We define seed motifs as either strings (e.g. ``just got fired''), 2-grams (e.g. (``just'', ``hired'')) or regexes (e.g. (``\verb/(^|\W)looking[\w\s\d]* gig[\W]/'').

To select initial seed motifs, we used the list of initial motifs elaborated by \citet{antenucci2014using}. We also imposed extra requirements on additional motifs, such as the presence of first-person pronouns (e.g. ``I got fired'' for the ``Lost Job'' class), as we restricted the analysis to the author's own labor market situation. We also used adverbs such as ``just'' to take into account the temporal constraint for classes ``Lost Job'' and ``Is Hired''. For Mexican Spanish and Brazilian Portuguese motifs, we both translated the English motifs and asked native speakers to confirm the relevance of the translations and add new seeds (e.g. “chamba” is a Mexican Spanish slang word for “work”). We then ran a similar selection process. 

For each of the candidate seed motif, we computed specificity and frequency on the random set $R_e$. For each class $\chi$, we defined specificity for a given motif $M$ as the share of positives for class $\chi$ in a random sample of 20 tweets from $R_e$ that contain $M$. The frequency of motif $M$ is defined as the share of tweets in $R_e$ that contain $M$. 

In order to have motifs that are both frequent and specific enough, we defined the following selection rule: we only retained motifs that have a specificity of or over 1\% and for which the product of specificity and frequency is above $1.10^{-7}$.

In total , we evaluated a total of 54 seeds for the US, 101 for Mexico and 42 for Brazil. After evaluation, we retained 26 seeds for the US, 26 for MX and 21 for Brazil. We report the retained motifs in Table \ref{tab:initial_motifs}. 

\subsection{Data labeling}
\label{appendix_data_labeling}

To label unemployment-related tweets, we used the crowdsourcing platform Amazon Mechanical Turk. This platform has the advantage of having an international workforce speaking several languages, including Spanish and Brazilian Portuguese on top of English.

For each tweet to label, turkers were asked the five questions listed in Table~\ref{tab:questions}.  Each turker was presented with a list of 50 tweets and each labeled tweet was evaluated by at least two turkers. 
A turker could choose to answer either \textit{yes}, \textit{no} or, \textit{I am not sure}. 
We included two attention check questions to exclude low-quality answers. Regarding the attention checks, we had the two following sentences labeled: ``I lost my job today'', which is a positive for class ``Lost Job'' and ``Is Unemployed'' and negative for the other classes, and ``I got hired today'', which is a positive for the class ``Is Hired'' and a negative for the other classes. We discarded answers of workers who didn't give the five correct labels for each quality check. To create a label for a given tweet, we required that at least two workers provided the same answer. A \textit{yes} was then converted to a positive label, a \textit{no} to a negative label, a tweet labeled by two workers as \textit{unsure} was dropped from the sample.

During this labeling process, all workers were paid with an hourly income above the minimum wage in their respective countries. For a labeling task of approximately 15 minutes, turkers from the US, Mexico and Brazil received respectively 5USD, 5USD and 3USD.

\input{tables/table_questions}

\subsection{Dataset description}
\label{appendix_dataset_description}

\subsubsection{Share of positives per class}

We provide descriptive statistics on the share of positives per class in the stratified sample for each language in Table \ref{tab:stratified_sample_description}. 

\input{tables/table_dataset_description_iter0}

\subsubsection{Class co-occurence}

In this section, we provide an analysis of the extent to which each class is mutually exclusive. For this, we focus on the English initial stratified sample.

First, the classes “Is Unemployed”, “Lost Job” and “Job Search” are non-mutually exclusive in many cases. As expected, the class “Lost Job” is highly correlated with the class “Is Unemployed” with 95\% of Lost Job positives being also positives for “Is Unemployed” in the US initial stratified sample (e.g. “i lost my job on monday so i'm hoping something would help.”, “as of today, for the first time in two years.....i am officially unemployed”). There are a few exceptions where users get hired quickly after being fired (e.g. “tfw you find a new job 11 days after getting laid off “). “Job Search” is also correlated with “Is Unemployed” (e.g. “I need a job, anyone hiring?”), though less than Lost Job, with 43\% of positives being also positives for “Is Unemployed” in the initial stratified sample. Cases where users are looking for a job but are not unemployed include looking for a second job (e.g. “need a second job asap.”) or looking for a better job while working (e.g. “tryna find a better job”). There are also a few ambiguous cases where users mention that they are looking for a job but it is not clear whether they are unemployed (e.g. “job hunting”) as well as edge cases where users just got hired but already are looking for another job (e.g. “i got hired at [company] but i don't like the environment any other suggestions for jobs ?”). For the class “Is Unemployed”, mutually exclusive examples are cases where the user only mentions her unemployment, without mentioning a recent job loss or the fact that she is looking for a job (e.g. “well i'm jobless so there's that”). 

Second, the classes “Is Hired” and “Job Offer” are essentially orthogonal from one another and from the other classes. The class “Is Hired” (e.g. “good morning all. started my new job yesterday. everyone was awesome.”) is almost always uncorrelated with the other classes apart from a few edge cases mentioned above. The class “Job Offer”  (e.g. “we are \#hiring process control/automation engineer job in atlanta, ga in atlanta, ga \#jobs \#atlanta”)  is almost always orthogonal to the other classes apart from a few exceptions. For instance, it can happen that a user who just got hired mentions job offers in her new company (e.g. “if you guys haven't been to a place called top golf i suggest you to go there or apply they are literally the best people ever i'm so happy i got hired”).

We detail the class co-occurrence in the US initial stratified sample in Table \ref{tab:class_cooccurence}. 
\input{tables/table_class_cooccurence}

\subsubsection{Additional descriptive statistics}

In this section, we include additional information about the US initial stratified sample. Table \ref{tab:avg_length_popular_words} contains information on average character length and most frequent tokens per class. Table \ref{tab:pos_tag_positives_stratified} describes the Part-of-speech tag distribution in positives across classes.

\input{tables/table_avg_length_popular_words}

\input{tables/table_pos_tags_positives_stratified}

\subsection{Pre-trained language model characteristics}
\label{appendix_model_characteristics}

To classify tweets in different languages and as mentioned in Section~\ref{finetuning_evaluation}, we used the following pre-trained language models from the Hugging Face model hub \cite{wolf2019transformers}:

\textbf{-- Conversational BERT\footnote{Available at \url{https://huggingface.co/DeepPavlov/bert-base-cased-conversational}}} for English tweets, trained and released by Deep Pavlov~\cite{burtsev2018deeppavlov}. This model was initialized with BERT base cased weights and shares the same configuration. It was then further pre-trained using a masked language modeling objective on an English corpus containing social media data (Twitter and Reddit), dialogues~\cite{li2017dailydialog}, debate transcripts~\cite{zhang2016conversational}, movie subtitles~\cite{lison2016opensubtitles2016} as well as blog posts~\cite{schler2006effects}. 

\noindent\textbf{-- BETO} for Spanish tweets~\cite{CaneteCFP2020}. This model has a BERT-base architecture and was pre-trained from scratch on a Spanish corpus derived from Wikipedia and the Spanish part of the OPUS project~\cite{tiedemann2012parallel}. 

\noindent\textbf{-- BERTimbau} for Brazilian Portuguese tweets~\cite{souza2020bertimbau}. This model also has a BERT-base architecture and was pre-trained from scratch on a large multi-domain Brazilian Portuguese corpus called brWaC~\cite{wagner2018brwac}.

All three language models have 110 million parameters. 

When it comes to the choice of language models for each language, the emerging literature considering language model pre-training on tweets to improve downstream tasks in the Twitter context gave us several potential candidates for English tweet classification. On top of Conversational BERT, we experimented with BERTweet \cite{nguyen2020bertweet}, which is the leader on the TweetEval leaderboard\footnote{The current leaderboard can be found here: \url{https://github.com/cardiffnlp/tweeteval}} as of March 2022 \cite{barbieri-etal-2020-tweeteval}. We also tested the performance of renowned pre-trained language models such as BERT base and RoBERTa base. We found that both Conversational BERT and BERTweet outperformed these well-known models for our task. Also, while BERTweet usually slightly outperformed Conversational BERT on the test set from the stratified sample in terms of AUROC, it had a worse performance on the random set $R_e$. This is why we chose Conversational BERT for English tweets.

For Spanish and Brazilian Portuguese tweets, in the absence of Twitter-specialized language models, we opted for the best performing pre-trained language models as of Fall 2020 for these languages, namely BETO for Spanish and BERTimbau for Brazilian Portuguese. We also experimented with multilingual language models, such as XLM-RoBERTa \cite{conneau-etal-2020-unsupervised}, but the monolingual approaches for Spanish and Brazilian Portuguese were performing better, both on the test set from the stratified sample and on the random set. 

\subsection{Fine-tuning and evaluation}
\label{appendix_finetuning_evaluation}

As mentioned in \ref{finetuning_evaluation} and following \citet{dodge2020fine}, we fine-tuned each BERT-based model with 15 different seeds and for 20 epochs. We evaluated the models 10 times per epoch and use early stopping with a patience of 11. We used a training and evaluation batch size of 8. The best model is defined as the best performing model in terms of area under the ROC curve (AUROC) on the evaluation set, at or after the first epoch.

As described in Algorithm \ref{fig:bigpicture}, we then ran the inference of the best model on both random sets $R_e$ and $R_s$. To speed up this inference process, we converted the PyTorch models to ONNX. 

In terms of computing infrastructure, we used either V100 (32GB) or RTX8000 (48GB) GPUs for the fine-tuning and parallelize inference over 2000 CPU nodes. The average runtime for fine-tuning and evaluation on the one hand and inference on the other hand is respectively of 45 minutes and 3 hours.

\subsection{Performance at iteration 0}

We report detailed AUROC results on the test set from the stratified sample in Table \ref{table_auc_iter0}. 
\input{tables/table_auc_iter0}

\subsection{Evaluation metrics}

In this section, we detail the evaluation process. The values of each metric across iterations for each language and each method can respectively be found in Table \ref{tab:avg_precision_cross_country} and \ref{tab:avg_precision_cross_method}.

\begin{figure*}[h]
    \centering
    \includegraphics[width=0.5\textwidth]{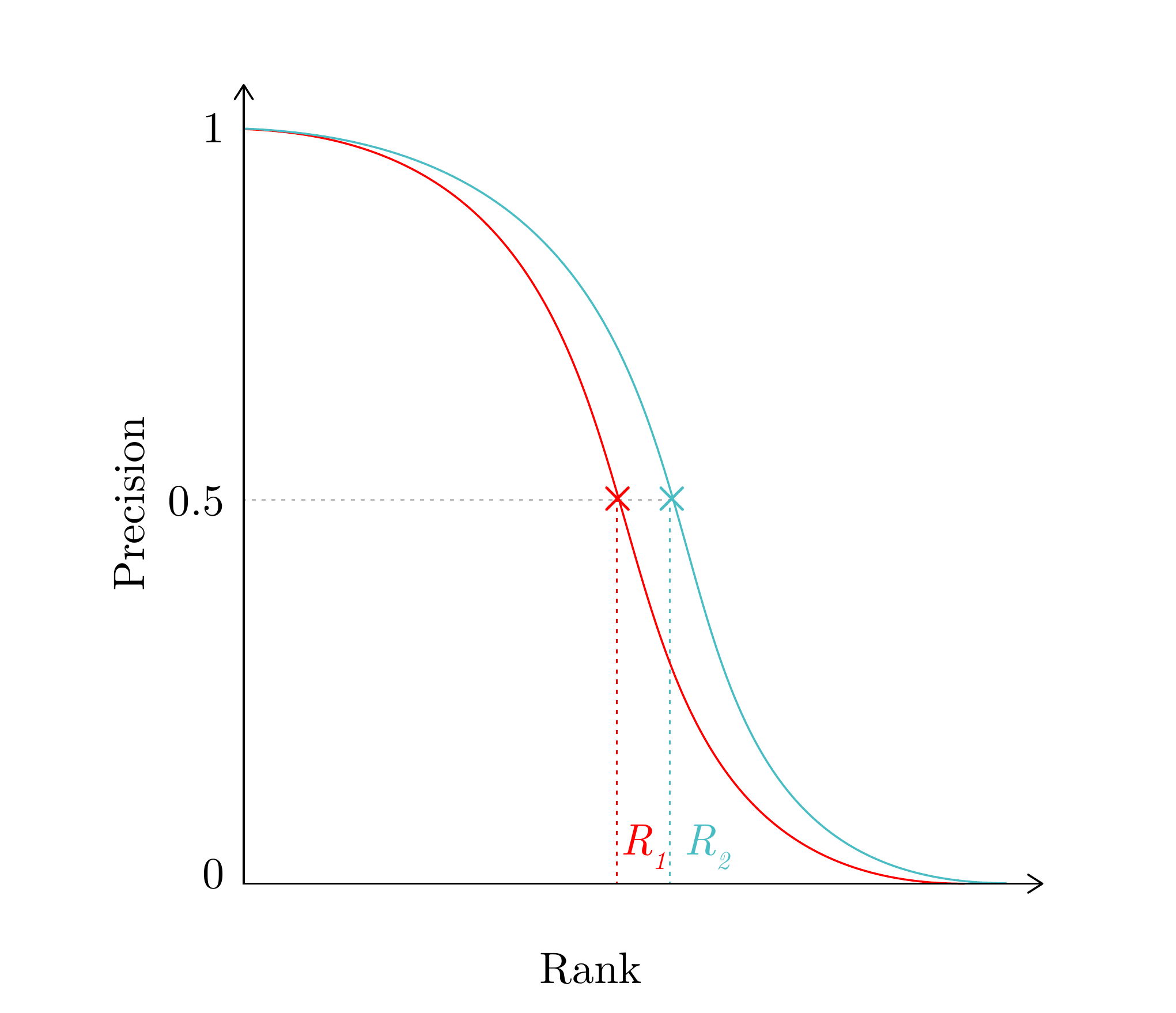} 
    \caption{\footnotesize{Illustration of the procedure used to determine the number of predicted positives. In this example, the number of predicted positives is $R_1$ for iteration 1 and $R_2$ for iteration 2.}
    \label{fig:clarification_eval_metrics}}
\end{figure*}

\label{appendix_evaluation_metrics}

\subsubsection{Sampling for evaluation}
\label{appendix_evaluation_sampling}

As mentioned in Section~\ref{evaluation_metrics}, for each country, AL strategy, iteration and class, we labeled 200 tweets along the BERT confidence score distribution. This tweet selection overweighted the top of the score distribution. Specifically, we retained tweets with the following ranks in the score distribution: 1-20; 101-110; 317-326; 1,001-1,010; 2,155-2,164; 4,642-4,651; 10,001-10,010; 17,783-17,792; 31,623-31,632; 56,235-56,244; 100,001-100,010; 158,490-158,499; 251,189-251,198; 398,108-398,117; 630,958-630,967; 1,000,001-1,000,010.



\subsubsection{Average Precision}

With the retained tweets, we computed the Average Precision (AP) at each iteration and for each class and language. We used the standard definition of AP in information retrieval and defined AP at iteration $i$ for class $c$ and method $m$ as: 

\[ AP_{i, c, m} = \frac{\sum_{r \in R_{i, c, m}}^{} P(r) \times pos(r)}{N_{i, c, m}} \]

where:

\begin{itemize}
    \item $R_{i, c, m}$ is the ensemble of ranks in the confidence score distribution of class $c$ at iteration $i$ and for method $m$ of all tweets sampled for evaluation and labeled for class $c$ and method $m$ both at iteration $i$ and preceding iterations
    \item $P(r)$ is the share of positives in sampled tweets with rank at iteration $i$ and for class $c$ inferior or equal to $r$
    \item $pos(r)$ is equal to 1 if tweet ranked $r$ for iteration $i$ and class $c$ is positive and 0 otherwise
    \item $N_{i,c, m}$ is the number of tweets sampled and labeled for class $c$ and method $m$ both at iteration $i$ and preceding iterations
\end{itemize} 

\subsubsection{Number of predicted positives}
\label{nb_predicted_positives}

We defined the number of predicted positives $E$ as the average rank in the confidence score distribution when the share of positives reaches 0.5. 
In practice, for each iteration $i$ and class $c$ and the related BERT model $M$, we first ranked the evaluation set $R_{e}$ according the prediction scores from $M$. We then binned the evaluation labels of each iteration until $i$ into 20 bins of equal size, and we estimated the proportion of positives in each bin and the average rank of each bin. We then identified the first bin for which the proportion of positive labels reaches 0.5. We estimated an upper and a lower bound for $E$ by taking the average rank of tweets included in the bin above and below the 0.5 cutoff respectively, and we estimated $E$ as the midpoint between its lower bound and its upper bound estimate. For each round, we report $E$ as well as its lower and upper bound estimates. We provide an illustration of this procedure in Figure \ref{fig:clarification_eval_metrics}. 

By convention, the number of predicted positives is equal to 1 when the proportion of positive labels sampled from the evaluation set remains below 0.5 for all ranks.

\subsubsection{Diversity of true positives}
\label{appendix_diversity}

To compute diversity for a given iteration $i$ and class $c$, we first encoded all positive tweets sampled for the evaluation of class $c$ at iteration $i$ as well as preceding iterations into sentence embeddings \cite{reimers-2019-sentence-bert}. To do so, we used the ``all-mpnet-base-v2'' model for English and the 
``paraphrase-multilingual-mpnet-base-v2'' model  for Spanish and Portuguese \cite{reimers-gurevych-2020-making}. These models are in open source access on the sentence-transformers GitHub repository\footnote{\url{https://github.com/UKPLab/sentence-transformers}}.

After computing the embeddings, we defined the diversity rate in a set of positive tweets as the mean pairwise distance between all possible pairs in this set. The pairwise distance between tweet A and B is defined as $1- sim(E_A, E_B)$ where $sim$ is a cosine similarity function and $E_A$ and $E_B$ are the sentence embeddings for tweets $A$ and $B$. By convention, diversity is equal to 0 when there is no more than 1 positive label.

\subsubsection{Standard error computation}

For average precision and diversity, we derived standard errors by using bootstrap samples on the pool of $N$ tweets used to compute the metric. We sampled with replacement $N$ tweets in this pool and repeated the process 1000 times. We then computed the metric for each of these samples and finally computed the mean and the standard error.

For the number of predicted positives, our method does not allow to directly use bootstrap. We therefore computed the upper and lower bound as described in Section \ref{nb_predicted_positives}.

\subsubsection{Convergence}
\label{convergence}

As stated in Section \ref{description_al_methods}, we considered that an AL strategy had converged when there was no significant variation of average precision, number of predicted positives and diversity for at least two iterations. 

To determine whether there is a significant variation in average precision and diversity from one iteration to the next, we performed t-tests. For the number of predicted positives, since we could only estimate an upper and lower bound, we considered that there was no significant variation from one iteration to the next if the interval between the lower bound and the upper bound overlapped from one iteration to the next.

We report in bold the metric values at convergence in Table \ref{tab:avg_precision_cross_country} and \ref{tab:avg_precision_cross_method}.

\subsection{Exploit-explore retrieval algorithm}
\label{appendix_algorithm}

In this section, we detail the functioning of the new AL strategy we coin exploit-explore retrieval in Algorithm \ref{algorithm_exploit_explore}. 
\input{tables/algorithm_exploit_explore}

We define the k-skip-n-grams used in this approach as follows: for a given text sequence $T$, the set of k-skip-n-grams, with $k$ a positive integer and $n$ in $\{2;3\}$, is made of all the ordered combinations of $n$ words in $T$. For instance, for $T =$ ``I am very happy'',  the set of k-skip-2grams is: \{ (I, am), (I, very), (I, happy), (am, very), (am, happy), (very, happy)\}. The $k$ blanks do not need to be successive. To define the k-skip-n-grams contained in tweets, each tweet was tokenized using the \textit{ekphrasis} package ~\cite{baziotis-pelekis-doulkeridis:2017:SemEval2}.

To decide on the $10^4$ threshold for top tweets, we estimated the base rate for each class and country. We defined the base rate for a given class as the share of positives for this class in the whole sample of tweets. To estimate this base rate for each class and country, we computed the specificity and frequency of each initial motif (listed in Table \ref{tab:initial_motifs}) and defined the base rate estimate as the sum over each motif of the motif's frequency weighted by its specificity. We detail the estimation results in Table \ref{tab:base_rates}.

The base ranks in our random sample of 100 million tweets $R_e$ (ie: base rate multiplied by $10^{8}$) ranged from $10^{2}$ to $10^{5}$ with a majority below $10^{4}$ in Mexico and Brazil. We tried $T=10^{3}$, $T=10^{4}$ and $T=10^{5}$ as candidate thresholds for the top tweets and they gave very similar results for the k-skip-n-grams used in the exploration step. We finally chose $10^{4}$ to balance between higher base ranks in the US and lower base ranks elsewhere. Our choice for the other hyperparameters were dictated by our budget constraint.

\input{tables/table_base_rates}

For illustration of the exploration part of this method, we detail the top-lift k-skip-n-grams selected from US tweets, for each iteration and for each class, in Table \ref{tab:emerging_motifs_US}.

\input{tables/table_emerging_motifs_US}
\subsection{Calibration for uncertainty sampling}
\label{appendix_calibration}

In order to calibrate the BERT confidence scores to do uncertainty sampling, we proceeded in the following way. 

For each country, AL strategy and class, we used the 200 tweets we retained along the confidence score distribution on $R_s$ and labeled for evaluation. From this labeled set, we built 10.000 balanced bootstrap samples and fit a logistic regression to each of these samples. We therefore obtained a set of 10.000 logistic regression parameter pairs $((\beta_{0, i}, \beta_{1, i}))_{i \in [1, 10.000]}$. 
We then used this set of parameters to find the BERT confidence score $x^{*}$ for which its calibrated version is equal to 0.5. To do so, we used Brent's method \cite{brent1971algorithm} and defined $x^{*}$ as the root of the following function:

$$ \frac{\sum_{i=1}^{10.000} \sigma(\beta_{0,i} + \beta_{1,i} x)}{10.000} - 0.5$$

where $\sigma$ is a standard logistic function. 

Knowing $x^{*}$, we were then able to perform uncertainty sampling by sampling tweets with confidence scores around $x^{*}$. 

\subsection{Additional experimental results}
\label{appendix_additional_results_cross_country}

\begin{figure*}[!ht]
    \centering
    \includegraphics[width=1\textwidth]{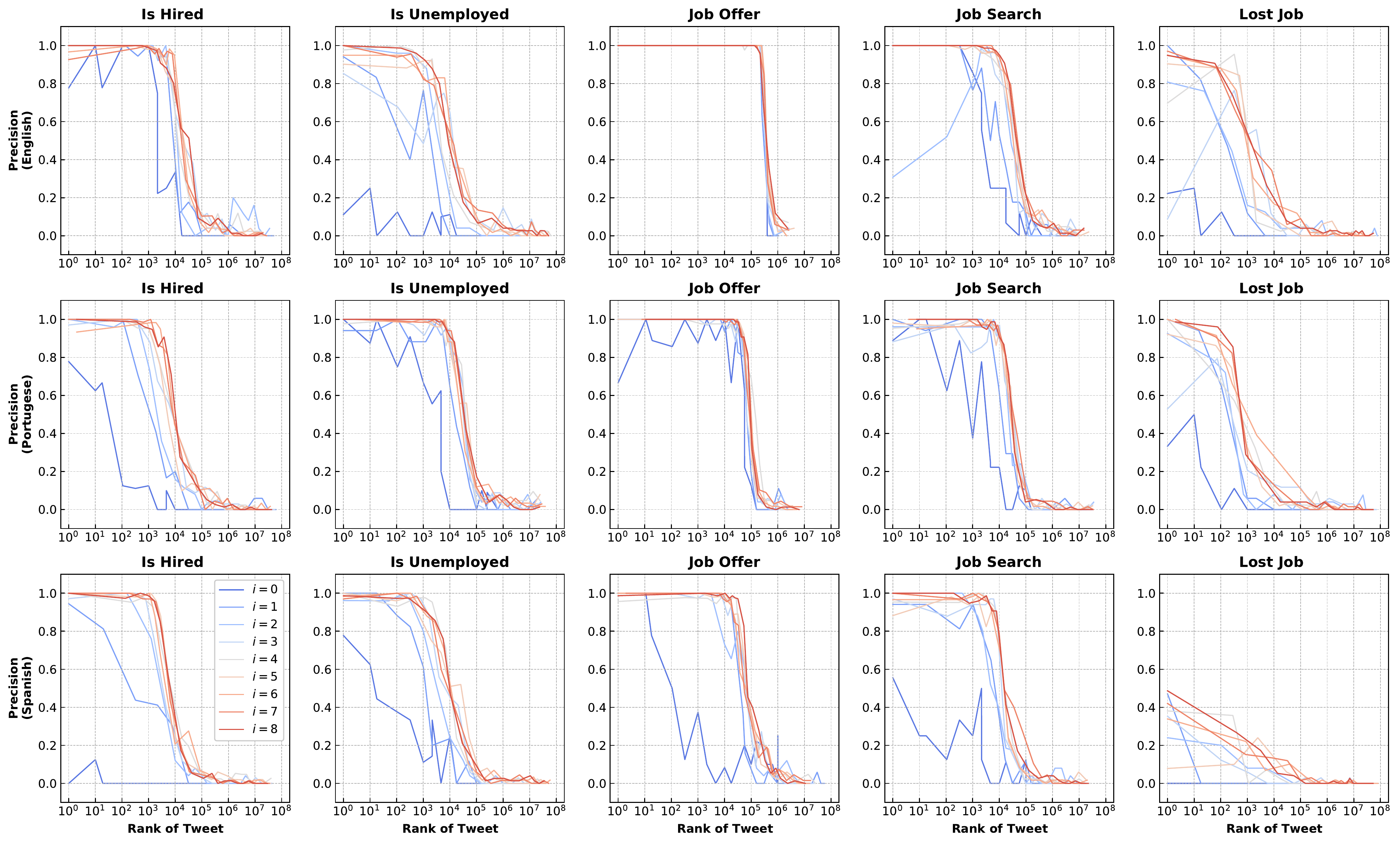} 
    \caption{\footnotesize{Precision (y-axis) as a function of tweet rank based on confidence score (i.e. positive label probability output by the model) (x-axis). For each language (in row) and class (in column), we ranked the tweets from the evaluation random set ${R_e}$ by their confidence score assigned by the BERT-based classifiers in descending order. We then sampled tweets along the rank distribution and labeled them. Each marker corresponds to a sample of 10 labeled tweets. Colors encode successive iterations of AL from 0 (blue) to 8 (red).}
    \label{fig:precision_country_comparison}}
\end{figure*} 

\begin{figure*}[!ht]
    \centering
    \includegraphics[width=1\textwidth]{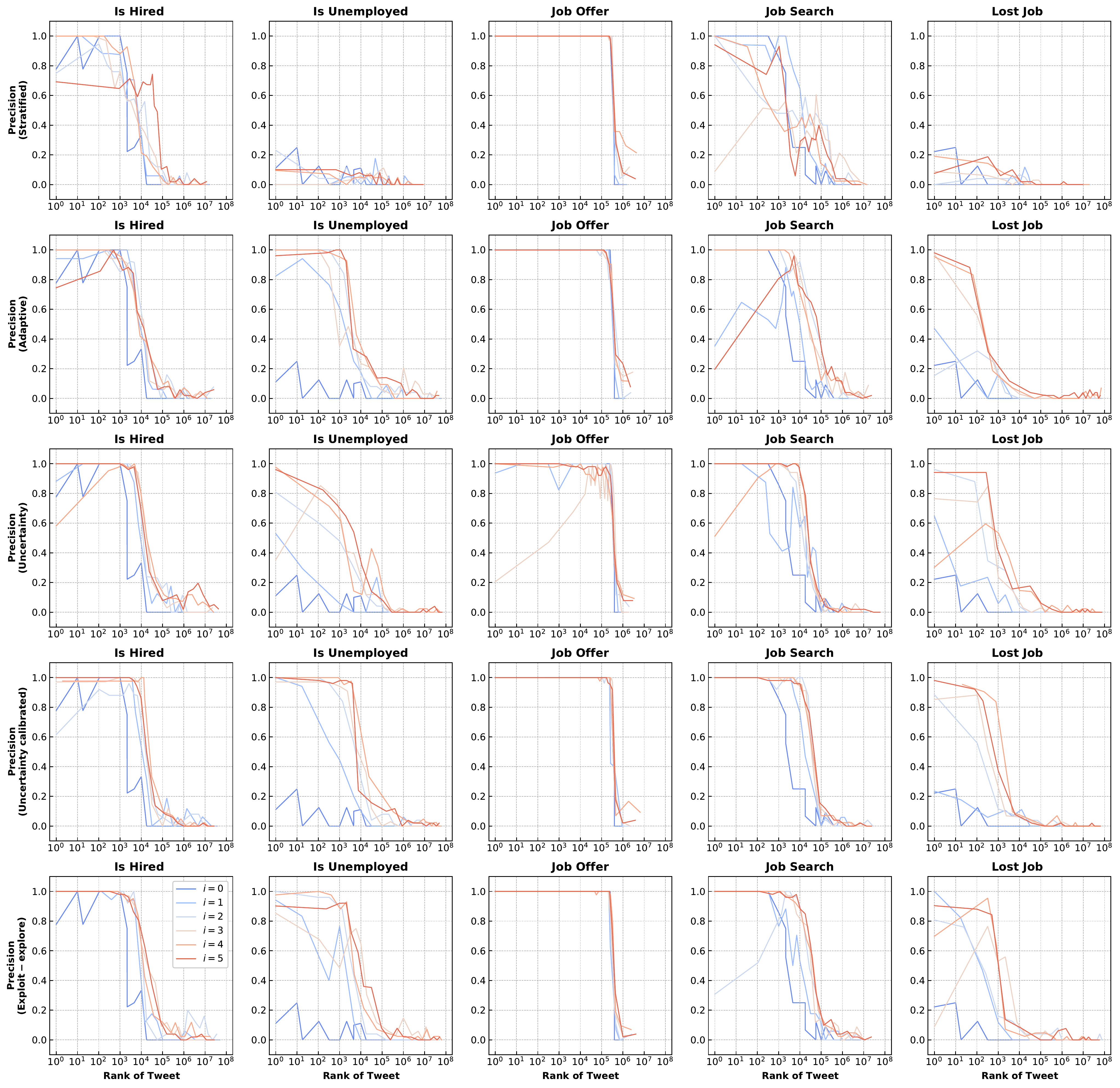} 
    \caption{\footnotesize{Precision (y-axis) as a function of tweet rank based on confidence score (i.e. positive label probability output by the model (x-axis)). For each AL strategy (in row) and class (in column), we ran the same process as the one described in Figure \ref{fig:precision_country_comparison}. Colors encode successive iterations of AL from 0 (blue) to 5 (red).}}
    \label{fig:precision_al_method_comparison}
\end{figure*} 




In this section, we report additional experimental results on precision and average precision. 

We report precision for the exploit-explore retrieval strategy across countries in Figure \ref{fig:precision_country_comparison} and for the four AL stategies on English tweets in Figure \ref{fig:precision_al_method_comparison}.

Also, we detail the evaluation results for the exploit-explore retrieval strategy across countries in Table \ref{tab:avg_precision_cross_country} and for the four AL stategies on English tweets in Figure \ref{tab:avg_precision_cross_method}.

\clearpage
\input{tables/table_avg_precision_cross_country}

\clearpage
\input{tables/table_avg_precision_cross_method}







\end{document}

%% file: tables/algorithm_experimental_procedure.tex
\setlength{\intextsep}{5pt} 
 \setlength{\floatsep}{0pt}
 \setlength{\textfloatsep}{0pt}
\begin{algorithm}[htbp]
  \textbf{Initialization}: for each seed $s$, sample 150 tweets containing $s$ from $R_s$; have them labeled for the five classes; the resulting labeled set is the stratified sample $L_{0} = S_0$; discard already sampled tweets from $R_s (R_s = R_s - L_0)$ \\
  \textbf{At each iteration $i$ and for each class:}
  \begin{itemize}[leftmargin=*, noitemsep, topsep=0pt]
      \item \textbf{Finetuning}: train-test split of $S_i$ in 70/30; finetune 15 BERT models on the train \\ set using different seeds; select the best \\ model $M_{i}^{*}$ with the highest AUROC on the \\ test set.
      \item \textbf{Inference on ${R_e}$ and $R_{s}$ using $M_{i}^{*}$}
      
      \item \textbf{Active Learning}: sample most informative tweets from $R_s$ (100 per class); have them labeled for the five classes; the resulting \\ labeled set is $L_{i+1}$; define $S_{i+1} = \bigcup_{j=0}^{i+1} L_{j}$ \\ and $R_s = R_s - L_{i+1}$
      
      \item \textbf{Evaluation:} sample tweets along the score distribution in $R_e$; have them labeled; \\ compute the average precision, number of predicted positives and \\ diversity metrics
  \end{itemize}
\caption{Experimental procedure}
\label{fig:bigpicture}
\end{algorithm}

%% file: tables/table_motifs.tex
\begin{table*}[ht!]

\centering
\scalebox{0.7}{
\begin{tabular}{c|l|l|l|l|l|l|l|l|l} 
\toprule
\textbf{Class} & \textbf{English motifs} & S\textsubscript{EN} & F\textsubscript{EN}  & \textbf{Spanish motifs} & S\textsubscript{SP} & F\textsubscript{SP} & \textbf{Portuguese motifs} & S\textsubscript{PT} & F\textsubscript{PT}  \\
\midrule
\parbox[t]{2mm}{\multirow{6}{*}{\rotatebox[origin=c]{90}{Is Unemployed}}} 
& \tabitem (i, unemployed) & 0.45 & 9.6e-6 & \tabitem estoy desempleado & 0.75 & 1.6e-6 & \tabitem estou desempregad[o/a] & 0.65 & 6e-6 \\ 
& \tabitem unemployed & 0.15 & 7.4e-5 & \tabitem sin empleo & 0.05 & 1.4e-5 & \tabitem (eu, sem, emprego) & 0.15 & 3.6e-6 \\ 
& \tabitem (i, jobless) & 0.45 & 2.4e-6 & \tabitem sin chamba & 0.15 & 1e-5 &  & & \\ 
& \tabitem jobless & 0.15 & 3.2e-5 & \tabitem nini & 0.15 & 4.9e-4 & & &\\ 
& \tabitem unemployment & 0.1 & 9e-5 & \tabitem no tengo trabajo/ & 0.5 & 8.6e-6 & & & \\
&  & & & chamba/empleo & & & & & \\ 
\midrule
\parbox[t]{2mm}{\multirow{7}{*}{\rotatebox[origin=c]{90}{Lost Job}}} 
& \tabitem (i, fired) & 0.05 & 4.9e-5 & \tabitem me despidieron & 0.2 & 2.6e-6 & \tabitem (perdi, emprego) & 0.35 & 3e-6 \\ 
& \tabitem i got fired & 0.25 & 3.3e-6 & \tabitem perdí mi trabajo &  0.2 & 5.3e-7 & \tabitem (perdi, trampo) & 0.15 & 1.6e-6 \\ 
& \tabitem just got fired & 0.2 & 2e-6 & \tabitem me corrieron & 0.1 & 1.1e-5 & \tabitem  fui demitido & 0.75 & 2.9e-6 \\ 
& \tabitem laid off & 0.2 & 1.2e-5 & \tabitem me quedé sin trabajo/ & 0.4 & 2.4e-6 & \tabitem me demitiram & 0.5 & 2.8e-7 \\ 
& \tabitem lost my job & 0.35 & 1.9e-6 & /chamba/empleo & & & \tabitem me mandaram embora & 0.25 & 6.7e-7 \\
&  & & & \tabitem ya no tengo trabajo/ & 0.55 & 9.8e-7 &  & & \\
&  & & & /chamba/empleo & & &  & & \\
\midrule
\parbox[t]{2mm}{\multirow{7}{*}{\rotatebox[origin=c]{90}{Job Search}}} 
& \tabitem (anyone, hiring) & 0.45 & 2e-6 & \tabitem (necesito, trabajo) & 0.7 & 2.5e-5 & \tabitem (gostaria, emprego) & 0.2 & 9.5e-7 \\ 
& \tabitem (wish, job) & 0.2 & 1.3e-5 & \tabitem (necesito, empleo) & 0.9 & 3.2e-6 & \tabitem (queria, emprego) & 0.45 & 1.5e-5 \\ 
& \tabitem (need, job) & 0.55 & 5.5e-5 &\tabitem (busco, trabajo) & 0.5 & 9e-6 & \tabitem (preciso, emprego) & 0.5 & 3.6e-5 \\ 
& \tabitem (searching, job) & 0.15 & 1.7e-6 & \tabitem (buscando, trabajo) & 0.45 & 1.7e-5 & \tabitem (procurando, emprego) & 0.25 & 1.5e-5 \\ 
& \tabitem (looking, gig) & 0.3 & 3.4e-6 & \tabitem (alguien, trabajo) & 0.1 & 3e-5 & & & \\ 
& \tabitem (applying, position) & 0.35 & 1.2e-6 & & & & & & \\ 
& \tabitem (find, job) & 0.3 & 8.9e-5 & & & & & & \\ 
\midrule
\parbox[t]{2mm}{\multirow{6}{*}{\rotatebox[origin=c]{90}{Is Hired}}} 
& \tabitem (found, job) & 0.25 & 6.2e-6 & \tabitem (conseguí, empleo) & 0.55 & 2.5e-5 & \tabitem (consegui, emprego) & 0.15 & 3e-5 \\ 
& \tabitem   (just, hired) & 0.15 & 9.4e-6 & \tabitem nuevo trabajo & 0.75 & 3.4e-5 & \tabitem fui contratad[o/a] & 0.45 & 2.6e-6 \\ 
& \tabitem  i got hired & 0.6 & 2e-6 & \tabitem nueva chamba & 0.45 & 3.3e-6 & \tabitem (começo, emprego) & 0.4 & 2.1e-6 \\ 
& \tabitem  (got, job) & 0.45 & 7.6e-5 & \tabitem (encontré, trabajo) & 0.25 & 4.7e-6 & \tabitem (novo, emprego/trampo) & 0.25 & 4.1e-5 \\ 
& \tabitem  new job & 0.25 & 8e-5 & \tabitem (empiezo, trabajar) & 0.4 & 4.5e-6 & \tabitem primeiro dia de trabalho & 0.65 & 1.3e-5 \\ 
&  & & & \tabitem primer día de trabajo & 0.55 & 2.3e-5 &  & & \\ 
\midrule 
\parbox[t]{2mm}{\multirow{5}{*}{\rotatebox[origin=c]{90}{Job Offer}}} 
& \tabitem job & 0.1 & 3e-3 & \tabitem empleo & 0.15 & 8.6e-4 & \tabitem (enviar, curr[i/í]culo) & 0.65 & 1.4e-5  \\ 
& \tabitem hiring & 0.2 & 5e-4 & \tabitem contratando & 0.35 & 2.9e-5 & \tabitem (envie, curr[i/í]culo) & 0.7 & 8e-6 \\ 
& \tabitem opportunity & 0.4 & 9.6e-4 & \tabitem empleo nuevo & 0.55 & 8.8e-7 & \tabitem (oportunidade, emprego) & 0.5 & 1.6e-5  \\
& \tabitem apply & 0.15 & 6.7e-4 & \tabitem vacante & 0.55 & 2e-4 & \tabitem (temos, vagas) & 0.45 & 1.5e-5 \\ 
&  & & & \tabitem estamos contratando & 0.9 & 9.7e-6 &  & & \\
\bottomrule 
\end{tabular}
}
\caption{Initial motifs for each language and class. The use of parentheses indicate regexes matching all strings containing the words in the parentheses in the order in which they are indicated. A slash separating several words indicates that the regex will match any of the candidate words separated by slashes. For each motif $M$ in country $c$, $S_{c}$ and $F_{c}$ are respectively $M$'s specificity and frequency in the evaluation random sample $R_e$.}
\label{tab:initial_motifs}

\end{table*}

%% file: tables/table_questions.tex
\begin{table}[!htbp]
\centering
\scalebox{.75}{
\begin{tabular}{p{0.06\textwidth}|p{0.5\textwidth}} 
\toprule
\textbf{Class} & \textbf{Question}\\ \midrule
Is Unemployed  & Does the tweet indicate that the person who wrote the tweet is currently (at the time of tweeting) unemployed? For example, tweeting “Now I am unemployed”, or “I just quit my job” is likely to indicate that the person who tweeted is currently unemployed. \\ 
\hline
Lost Job       & Does this tweet indicate that the person who wrote the tweet became unemployed within the last month? For example, tweeting “I lost my job today”, or “I was fired earlier this week” is likely to indicate that the person who tweeted became unemployed within the last month. \\ 
\hline
Job Search     & Does this tweet indicate that the person who wrote the tweet is currently searching for a job? For example, tweeting “I am looking for a job”, or “I am searching for a new position” is likely to indicate that the person who tweeted is currently searching for a job. \\ 
\hline
Is Hired       & Does this tweet indicate that the person who wrote the tweet was hired within the last month? For example, tweeting “I just found a job”, or “I got hired today” is likely to indicate that the person who tweeted was hired within the last month. \\ 
\hline
Job Offer      & Does this tweet contain a job offer? For example, tweeting “Looking for a new position?”, or “Here is a job opportunity you might be interested in” is likely to indicate that the tweet contains a job offer.\\
\bottomrule
\end{tabular}
}
\caption{List of questions asked to the Amazon Turkers when labelling each tweet}
\label{tab:questions}

\end{table}

%% file: tables/table_dataset_description_iter0.tex
\begin{table*}[htbp]
\centering
\begin{tabular}{c|c|c|c|c|c|c}
\toprule
\multirow{2}{*}{\textbf{Language}} & \multirow{2}{*}{\textbf{Label}} & \multicolumn{5}{c}{\textbf{Class}}                                   \\ \cline{3-7} 
                         &                        & Lost Job & Is Hired & Is Unemployed & Job Offer & Job Search \\ \midrule
\multirow{3}{*}{English}      & yes                    & 270      & 334      & 796           & 600       & 524        \\ \cline{2-7} 
                         & no                     & 4239     & 4181     & 3710          & 3918      & 3993       \\ \cline{2-7} 
                         & unsure                 & 15       & 9        & 18            & 6         & 7          \\ \hline
\multirow{3}{*}{Spanish}  & yes                    & 213      & 388      & 1116          & 515       & 659        \\ \cline{2-7} 
                         & no                     & 3488     & 3331     & 2579          & 3210      & 3059       \\ \cline{2-7} 
                         & unsure                 & 28       & 10       & 34            & 4         & 11         \\ \hline
\multirow{3}{*}{Portuguese}  & yes                    & 175      & 422      & 925           & 485       & 614        \\ \cline{2-7} 
                         & no                     & 2514     & 2272     & 1761          & 2215      & 2084       \\ \cline{2-7} 
                         & unsure                 & 14       & 9        & 17            & 3         & 5          \\ \bottomrule
\end{tabular}
\caption{Label distribution on the stratified sample for each country and class}
\label{tab:stratified_sample_description}
\end{table*}

%% file: tables/table_class_cooccurence.tex
\begin{table*}[!htb]
\centering
\begin{tabular}{l|lllll}
\toprule
\multirow{2}{*}{\textbf{Class}} & \multicolumn{5}{c}{\textbf{Share of positives per class (in \%)}}                                                       \\ \cline{2-6} 
 & \multicolumn{1}{l|}{Is Unemployed} & \multicolumn{1}{l|}{Lost Job} & \multicolumn{1}{l|}{Job Search} & \multicolumn{1}{l|}{Is Hired} & Job Offer \\ \midrule
Is Unemployed          & \multicolumn{1}{l|}{100} & \multicolumn{1}{l|}{32}  & \multicolumn{1}{l|}{28}  & \multicolumn{1}{l|}{1.3} & 0   \\ \hline
Lost Job               & \multicolumn{1}{l|}{95}  & \multicolumn{1}{l|}{100} & \multicolumn{1}{l|}{10}  & \multicolumn{1}{l|}{4}   & 0   \\ \hline
Job Search             & \multicolumn{1}{l|}{43}  & \multicolumn{1}{l|}{5}   & \multicolumn{1}{l|}{100} & \multicolumn{1}{l|}{2}   & 0   \\ \hline
Is Hired               & \multicolumn{1}{l|}{3.2} & \multicolumn{1}{l|}{3.2} & \multicolumn{1}{l|}{2.9} & \multicolumn{1}{l|}{100} & 2.3 \\ \hline
Job Offer              & \multicolumn{1}{l|}{0}   & \multicolumn{1}{l|}{0}   & \multicolumn{1}{l|}{0}   & \multicolumn{1}{l|}{1.3} & 100 \\ \bottomrule
\end{tabular}
\caption{Class co-occurrence in the US initial stratified sample. It reads as follows: out of all positives for the Is Unemployed class, 32\% are positives for Lost Job.}
\label{tab:class_cooccurence}

\end{table*}

%% file: tables/table_avg_length_popular_words.tex
\begin{table*}[!htb]
\centering
\resizebox{\textwidth}{!}{\begin{tabular}{l|l|r|r|r|r|r|r|r|r|r|r}
\toprule
\multirow{2}{*}{\textbf{Class}} &
  \multirow{2}{*}{\textbf{Average length}} &
  \multicolumn{10}{c}{\textbf{Top 10 most common tokens}} \\ \cline{3-12} 
 &
   &
  1 & 2 & 3 & 4 & 5 & 6 & 7 & 8 & 9 & 10 \\ \midrule
Is Unemployed &
  105 &
  i &
  job &
  a &
  to &
  my &
  and &
  the &
  for &
  fired &
  got \\ \hline
Lost Job &
  103 &
  i &
  my &
  got &
  fired &
  job &
  just &
  a &
  to &
  and &
  the \\ \hline
Job Search &
  96 &
  i &
  a &
  job &
  for &
  to &
  the &
  anyone &
  and &
  hiring &
  in \\ \hline
Is Hired &
  99 &
  i &
  job &
  a &
  got &
  the &
  my &
  and &
  new &
  hired &
  to \\ \hline
Job Offer &
  128 &
  job &
  a &
  for &
  in &
  jobs &
  hiring &
  to &
  at &
  the &
  \#\#q \\ \bottomrule
\end{tabular}}
\caption{Average character length and top 10 most frequent tokens for each class in the initial US stratified sample}
\label{tab:avg_length_popular_words}
\end{table*}

%% file: tables/table_pos_tags_positives_stratified.tex
\begin{table*}[!htb]
\centering
\begin{tabular}{c|ccccc}
\toprule
\multirow{2}{*}{\textbf{POS tag}} & \multicolumn{5}{c}{\textbf{Share per class (in \%)}}                                                                    \\ \cline{2-6} 
     & \multicolumn{1}{c|}{Is Unemployed} & \multicolumn{1}{c|}{Lost Job} & \multicolumn{1}{c|}{Job Search} & \multicolumn{1}{c|}{Is Hired} & Job Offer \\ \midrule
ADJ                               & \multicolumn{1}{c|}{7.18} & \multicolumn{1}{c|}{6.21}  & \multicolumn{1}{c|}{6.92}  & \multicolumn{1}{c|}{7.95}  & 8.15  \\ \hline
ADP                               & \multicolumn{1}{c|}{7.43} & \multicolumn{1}{c|}{7.61}  & \multicolumn{1}{c|}{8.27}  & \multicolumn{1}{c|}{7.67}  & 9.46  \\ \hline
ADV                               & \multicolumn{1}{c|}{6.85} & \multicolumn{1}{c|}{8.47}  & \multicolumn{1}{c|}{6.13}  & \multicolumn{1}{c|}{6.78}  & 3.75  \\ \hline
AUX                               & \multicolumn{1}{c|}{8.60} & \multicolumn{1}{c|}{9.52}  & \multicolumn{1}{c|}{6.96}  & \multicolumn{1}{c|}{7.86}  & 5.30  \\ \hline
CCONJ                             & \multicolumn{1}{c|}{4.23} & \multicolumn{1}{c|}{4.02}  & \multicolumn{1}{c|}{3.50}  & \multicolumn{1}{c|}{4.58}  & 2.89  \\ \hline
DET                               & \multicolumn{1}{c|}{6.71} & \multicolumn{1}{c|}{5.82}  & \multicolumn{1}{c|}{8.95}  & \multicolumn{1}{c|}{7.86}  & 6.81  \\ \hline
INTJ                              & \multicolumn{1}{c|}{1.85} & \multicolumn{1}{c|}{2.26}  & \multicolumn{1}{c|}{1.53}  & \multicolumn{1}{c|}{1.45}  & 0.72  \\ \hline
NOUN                              & \multicolumn{1}{c|}{9.73} & \multicolumn{1}{c|}{9.64}  & \multicolumn{1}{c|}{10.61} & \multicolumn{1}{c|}{10.22} & 11.00 \\ \hline
NUM                               & \multicolumn{1}{c|}{2.07} & \multicolumn{1}{c|}{2.07}  & \multicolumn{1}{c|}{1.66}  & \multicolumn{1}{c|}{2.24}  & 2.74  \\ \hline
PART                              & \multicolumn{1}{c|}{4.54} & \multicolumn{1}{c|}{4.06}  & \multicolumn{1}{c|}{3.96}  & \multicolumn{1}{c|}{3.85}  & 2.61  \\ \hline
PRON                              & \multicolumn{1}{c|}{9.97} & \multicolumn{1}{c|}{10.07} & \multicolumn{1}{c|}{9.90}  & \multicolumn{1}{c|}{9.40}  & 5.56  \\ \hline
PROPN                             & \multicolumn{1}{c|}{5.00} & \multicolumn{1}{c|}{5.19}  & \multicolumn{1}{c|}{4.31}  & \multicolumn{1}{c|}{5.96}  & 8.37  \\ \hline
PUNCT                             & \multicolumn{1}{c|}{8.54} & \multicolumn{1}{c|}{8.59}  & \multicolumn{1}{c|}{9.55}  & \multicolumn{1}{c|}{8.36}  & 10.39 \\ \hline
SCONJ                             & \multicolumn{1}{c|}{4.01} & \multicolumn{1}{c|}{3.01}  & \multicolumn{1}{c|}{3.69}  & \multicolumn{1}{c|}{2.40}  & 1.38  \\ \hline
SPACE                             & \multicolumn{1}{c|}{1.13} & \multicolumn{1}{c|}{1.05}  & \multicolumn{1}{c|}{1.04}  & \multicolumn{1}{c|}{1.14}  & 2.19  \\ \hline
SYM                               & \multicolumn{1}{c|}{1.27} & \multicolumn{1}{c|}{1.29}  & \multicolumn{1}{c|}{1.37}  & \multicolumn{1}{c|}{1.07}  & 5.89  \\ \hline
VERB & \multicolumn{1}{c|}{10.02}         & \multicolumn{1}{c|}{10.23}    & \multicolumn{1}{c|}{10.82}      & \multicolumn{1}{c|}{10.22}    & 10.28     \\ \hline
X                                 & \multicolumn{1}{c|}{0.85} & \multicolumn{1}{c|}{0.90}  & \multicolumn{1}{c|}{0.83}  & \multicolumn{1}{c|}{0.98}  & 2.52  \\ \bottomrule
\end{tabular}
\caption{Part-of-Speech (POS) tag distribution among positives of each class from the initial US stratified sample. The definition of the acronyms can be found \href{(https://universaldependencies.org/u/pos/}{here}.}
\label{tab:pos_tag_positives_stratified}
\end{table*}

%% file: tables/table_auc_iter0.tex
\begin{table*}[!htb]
\centering
\begin{tabular}{c|c|r|r|r|l|l}
\toprule
\multirow{2}{*}{\textbf{Language}} & \multirow{2}{*}{\textbf{Model}} & \multicolumn{5}{c}{\textbf{Class}}                                                                                                  \\ \cline{3-7} 
                         &                        & \multicolumn{1}{c|}{Lost Job} & \multicolumn{1}{c|}{Is Hired} & \multicolumn{1}{c|}{Is Unemployed} & Job Offer & Job Search \\ \midrule
English                       & Conversational BERT    &  0.959                         &  0.976                         & 0.965                              & 0.985     & 0.98       \\ \hline
Spanish                   & BETO                   & 0.944                         & 0.98                          & 0.949                              & 0.993     & 0.959      \\ \hline
Portuguese                   & BERTimbau              & 0.978                         & 0.973                         & 0.949                              & 0.991     & 0.971      \\ \bottomrule

\end{tabular}
\caption{AUROC results on the evaluation set at iteration 0.}
\label{table_auc_iter0}
\end{table*}

%% file: tables/algorithm_exploit_explore.tex
\begin{algorithm*}

 \textbf{Initialization:}
 
 $\forall k \in \mathbb{N}^{*}$ and n=2,3, determine all ordered k-skip-n-grams in the random set $R_s$ used to sample tweets for labelling. This results in a set of 2-grams $S_2$ and 3-grams $S_3$\;
 For n=2,3, discard all k-skip-n-grams from $S_n$ that:
 \begin{itemize}[leftmargin=*, noitemsep]
     \item contain one-grams made of at least one subtoken that is not in the BERT model vocabulary
     \item contain at least one repetition (e.g. (i, i, job))
     \item that have a frequency lower than 1 in 100K
 \end{itemize}
 \textbf{At each iteration $i$:}
 
 Discard tweets that were sampled and labeled at iteration $i-1$ from $R_s$ \;
 For each class $\chi$:
 \begin{itemize}
     \item Run inference on $R_s$ with the best BERT-based classifier for class $\chi$
     \item \textbf{Exploitation}: sample 50 tweets from the set of top 10,000 tweets in terms of confidence score assigned by the BERT-based classifier
    \item \textbf{Exploration}: for n=2,3,
    \begin{itemize}
        \item Compute lift for each k-skip-n-gram in $S_n$ 
        \item Discard all k-skip-n-grams from $S_n$ that 
        
        (1) were used to sample tweets for class $\chi$ at iteration $i-1$ and/or

        (2) have at least one one-gram in common with another k-skip-n-gram. 
        
        Only the k-skip-n-gram with the highest lift is kept. 
        \item Select 5 top-lift k-skip-n-grams in $S_n$
        \item For each retained top-lift k-skip-n-gram, sample 5 tweets in $R_s$ containing this motif
    \end{itemize}

 \end{itemize}
 Label sampled tweets for each class\;
 Add new sampled tweets to the set of all labels\;
 Perform new train-test split on this set and use this split to train and evaluate the classifier for the next iteration\;
 \caption{Exploit-explore retrieval}
\label{algorithm_exploit_explore}

\end{algorithm*}


%% file: tables/table_base_rates.tex
\begin{table*}[hbt!]
\centering

\begin{tabular}{l|l|l}
\toprule
\textbf{Language} & \textbf{Class} & \textbf{Base rate}  \\ \midrule
             English    &  Is Hired & $3.03 \times 10^{-4}$ \\ 
             English    &  Is Unemployed &  $2.16 \times 10^{-4}$ \\ 
             English    &  Job Offer & $5.38 \times 10^{-3}$  \\ 
             English    &  Job Search & $4.8 \times 10^{-4}$   \\ 
             English    &  Lost Job & $2.04 \times 10^{-5}$  \\ \midrule
             Spanish    &  Is Hired & $5.64\times 10^{-5}$ \\ 
             Spanish    &  Is Unemployed  & $8.16\times 10^{-5}$ \\ 
             Spanish    &  Job Offer & $2.58\times 10^{-4}$ \\ 
             Spanish    &  Job Search & $3.55\times 10^{-5}$ \\ 
             Spanish    &  Lost Job & $1.46\times 10^{-5}$ \\ \midrule
             Portuguese    &  Is Hired & $ 4.82 \times 10^{-5}$  \\ 
             Portuguese    &  Is Unemployed & $ 7.51 \times 10^{-5}$   \\ 
             Portuguese    &  Job Offer & $ 4.59 \times 10^{-5}$  \\ 
             Portuguese    &  Job Search &  $ 7.57 \times 10^{-5}$  \\ 
             Portuguese    &  Lost Job &  $ 3.91 \times 10^{-6}$ \\ \bottomrule
\end{tabular}
\caption{Estimated base rate for each country and class.}
\label{tab:base_rates}
\end{table*}

%% file: tables/table_emerging_motifs_US.tex
\begin{table*}[ht!]

\centering
\scalebox{0.7}{
\begin{tabular}{c|l|l|l|l} 
\toprule
\textbf{Class} & \textbf{Iteration 1} & \textbf{Iteration 2} & \textbf{Iteration 3} & \textbf{Iteration 4} \\
\midrule
\parbox[t]{2mm}{\multirow{10}{*}{\rotatebox[origin=c]{90}{Is Unemployed}}} 
& \tabitem (got, headache) & \tabitem (am, homeless) & \tabitem (got, fired) & \tabitem (am, clueless) \\ 
& \tabitem (having, breakdown) & \tabitem (lost, job) & \tabitem (in, desperately) & \tabitem (i, homeless) \\ 
& \tabitem (lost, voice) & \tabitem (need, broke) & \tabitem (job, hunting) & \tabitem (im, broke) \\ 
& \tabitem (im, losing) & \tabitem (unemployed, a) & \tabitem (laid, i) & \tabitem (in, limbo) \\ 
& \tabitem (m, depressed) & \tabitem (been, single, for) & \tabitem (unemployed, and) & \tabitem (unemployed, to)\\
& \tabitem (got, a, headache) & \tabitem (homeless, in, to) & \tabitem (got, fired, the) & \tabitem (i, am, homeless)\\
& \tabitem (am, having, attack) & \tabitem (i, am, unemployed) & \tabitem (have, no, life) & \tabitem (laid, off, and)\\
& \tabitem (i, lost, phone) & \tabitem (really, need, job) & \tabitem (laid, off, to) & \tabitem (m, broke, to\\
& \tabitem (im, in, need) &  & \tabitem (lost, my, up) & \tabitem (need, job ,can)\\
& \tabitem (losing, my, mind) &  & \tabitem (need, job, i) & \tabitem (strong, have, been)\\

\midrule
\parbox[t]{2mm}{\multirow{10}{*}{\rotatebox[origin=c]{90}{Lost Job}}} 
& \tabitem (broke, today) & \tabitem (fired, me) & \tabitem (been, sick) & \tabitem (just, fired) \\ 
& \tabitem (fell, bed) &\tabitem (got, laid) & \tabitem (fired, my) & \tabitem (now, pissed) \\ 
& \tabitem (got, hospital) & \tabitem (unfollowed, checked) & \tabitem  (got, banned) & \tabitem (today, sucked) \\
& \tabitem (just, kicked) & \tabitem ([, by, ]) & \tabitem (just, cancelled) & \tabitem (unemployed, for) \\ 
& \tabitem (lost, yesterday) & \tabitem (am, sick, again) & \tabitem (worked, days) & \tabitem (i, was, fired)\\
& \tabitem (got, kicked, of) & \tabitem (and, me, checked) & \tabitem (been, sick, for) & \tabitem (just, went, from)\\ 
& \tabitem (i, lost, today) &\tabitem (quit, my, job) & \tabitem (don, have, weekend) & \tabitem (my, job, today)\\ 
& \tabitem (just, pulled, over) & \tabitem (to, i, fired) & \tabitem  (i, fired, my) & \tabitem (now, am, pissed)\\ 
& \tabitem (out, the, hospital) &  & \tabitem (today, bad, day) & \\ 
& \tabitem (phone, last, night) &  &  & \\
\midrule
\parbox[t]{2mm}{\multirow{10}{*}{\rotatebox[origin=c]{90}{Job Search}}} 
& \tabitem (any, places) & \tabitem (any, jobs) & \tabitem (applying, i) & \tabitem (applying, for)\\ 
& \tabitem (job, asap) &\tabitem (interview, wish) & \tabitem (interview, tomorrow) & \tabitem (interview, get)\\ 
& \tabitem (know, hiring) & \tabitem (job, anyone) & \tabitem  (job, luck) & \tabitem (need, paying)\\ 
& \tabitem (need, second) & \tabitem (knows, let) & \tabitem (please, pls) & \tabitem (second, job)\\ 
& \tabitem (new, suggestions) & \tabitem (need, hiring) & \tabitem (anyone, knows, of) & \tabitem (that, hiring)\\
& \tabitem (if, anyone, knows) & \tabitem (a, second, job) & \tabitem (have, wish, luck) &  \tabitem (got, a, interview)\\ 
& \tabitem (am, for, jobs) &\tabitem (am, any, suggestions) & \tabitem (job, need, i) & \tabitem (hope, get, job)\\ 
& \tabitem (hiring, i, a) & \tabitem (got, an, interview) & \tabitem  (places, that, are) & \tabitem (i, need, second)\\ 
& \tabitem (need, new, job) & \tabitem (i, looking, anyone) & \tabitem (to, interview, me) & \\ 
& \tabitem (something, do, tonight) & \tabitem (knows, me, how) &  & \\ 

\midrule
\parbox[t]{2mm}{\multirow{10}{*}{\rotatebox[origin=c]{90}{Is Hired}}} 
& \tabitem (first, nervous) & \tabitem (got, accepted) & \tabitem (excited, job) & \tabitem (got, \$\$) \\ 
& \tabitem (got, hired) &\tabitem (hired, at) & \tabitem (hired, on) & \tabitem (hired, for)\\ 
& \tabitem (job, excited) & \tabitem (start, job) & \tabitem  (start, new) & \tabitem (i, promoted) \\ 
& \tabitem (new, woot) & \tabitem (started, weeks) & \tabitem (first, at, day) & \tabitem (job, tomorrow) \\ 
& \tabitem (start, tomorrow) & \tabitem (tomorrow, nervous) & \tabitem (hired, to, a) & \tabitem (first, day, new)\\ 
& \tabitem (finally, a, phone) & \tabitem (first, at, new) & \tabitem (i, job, got) & \tabitem (it, can, oh) \\
& \tabitem (i, hired, and) & \tabitem (job, i, got) & \tabitem (start, tomorrow, and) & \tabitem (just, call, from) \\ 
& \tabitem (start, my, new) & \tabitem (start, my, tomorrow) & \tabitem (starting, my, new) & \tabitem (start, my, job)\\ 
& \tabitem (the, job, got) & \tabitem (started, a, ago) &  & \\ 
& \tabitem (tomorrow, first, day) &  & \\ 

\midrule 
\parbox[t]{2mm}{\multirow{10}{*}{\rotatebox[origin=c]{90}{Job Offer}}} 
& \tabitem (apply, arc) & \tabitem (hiring, view) & \tabitem (apply, career) & \tabitem (apply, retail) \\ 
& \tabitem (click, jobs) &\tabitem (it, analyst) & \tabitem (click, job) & \tabitem (are, hospitality) \\ 
& \tabitem (recommend, career) & \tabitem (job, details) & \tabitem (hiring, hospitality) & \tabitem (click, arc)  \\ 
& \tabitem (anyone, retail) & \tabitem (manager, apply) & \tabitem (view, details) & \tabitem (job, circle) \\ 
& \tabitem (technician, hiring) & \tabitem (position, open) & \tabitem (we, arc) & \tabitem (needed, hiring) \\ 
& \tabitem (click, apply, job) & \tabitem (hiring, it, details) & \tabitem (hiring, to, career) & \tabitem (are, apply, career)\\
& \tabitem (now, developer, in) & \tabitem (is, apply, jobs) & \tabitem (it, view, details) & \tabitem (click, job, jobs) \\ 
& \tabitem (recommend, anyone, this) &\tabitem (job, analyst, view) & \tabitem (now, manager, in) & \tabitem (manager, new, york) \\ 
& \tabitem (we, jobs, career) & \tabitem (now, opportunities, in) & \tabitem  (we, apply, job) & \tabitem (now, hiring, circle) \\ 
& & \tabitem (we, are, assistant) &  & \tabitem (we, to, arc) \\ 

\bottomrule 
\end{tabular}
}
\caption{Top-lift k-skip-n-grams for each class and iteration of the Explore-Exploit Retrieval on US tweets. The fact that not all (class, iteration) pair have 10 k-skip-n-grams is explained by the fact that some set of tweets containing a top-lift k-skip-n-gram could not be labeled because of disagreement between crowdworkers on the right label to assign. }
\label{tab:emerging_motifs_US}

\end{table*}

%% file: tables/table_avg_precision_cross_country.tex
\begin{table*}[hbt!]
\centering
\small
\scalebox{0.7}{
\begin{tabular}{@{}cccccccccccc@{}}
\toprule
                         &                        &                         & i=0                                     & i=1                                     & i=2                                     & i=3                                              & i=4                                              & i=5                                              & i=6                                              & i=7                                              & i=8                                     \\ \midrule
\multicolumn{1}{c|}{IH} & \multicolumn{1}{c|}{P} & \multicolumn{1}{c|}{EN} & \multicolumn{1}{c|}{83.0 (3.8)}         & \multicolumn{1}{c|}{96.1 (0.7)}         & \multicolumn{1}{c|}{94.3 (1.1)}         & \multicolumn{1}{c|}{\textbf{96.3 (0.6)}}         & \multicolumn{1}{c|}{96.0 (0.7)}                  & \multicolumn{1}{c|}{96.5 (0.5)}                  & \multicolumn{1}{c|}{95.7 (0.9)}                  & \multicolumn{1}{c|}{93.8 (1.3)}                  & \multicolumn{1}{c}{96.2 (0.4)}         \\
\multicolumn{1}{c|}{IH} & \multicolumn{1}{c|}{P} & \multicolumn{1}{c|}{PT} & \multicolumn{1}{c|}{68.2 (7.6)}         & \multicolumn{1}{c|}{90.6 (1.7)}         & \multicolumn{1}{c|}{93.7 (1.2)}         & \multicolumn{1}{c|}{91.0 (1.8)}                  & \multicolumn{1}{c|}{93.9 (0.8)}                  & \multicolumn{1}{c|}{\textbf{94.9 (0.7)}}         & \multicolumn{1}{c|}{91.5 (1.6)}                  & \multicolumn{1}{c|}{95.8 (0.6)}                  & \multicolumn{1}{c}{95.3 (0.6)}         \\
\bottomrule
\end{tabular}
}
\caption{\footnotesize{Evaluation results using the exploit-explore retrieval active learning method. The results are reported across languages -- English ('EN'), Portugese ('PT'), Spanish ('ES') -- performance metrics -- average precision ('P'), number of predicted positives ('E'), diversity ('D') -- and classes -- is hired ('IH'), is unemployed ('IU'), job offer ('JO'), job search ('JS'), job loss ('LJ'). Standard errors for P and D are shown in parentheses, and we report a lower bound and an upper bound for E. Bold values indicate the iteration at which a model converges.}}
\label{tab:avg_precision_cross_country}
\end{table*}

%% file: tables/table_avg_precision_cross_method.tex
\begin{table*}[hbt!]
\centering
\small
\scalebox{0.7}{
\begin{tabular}{@{}ccccccccc@{}}
\toprule
                        &                        &                         & i=0                                     & i=1                                     & i=2                                     & i=3                                              & i=4                                              & i=5                                              \\ \midrule
\multicolumn{1}{c|}{IH} & \multicolumn{1}{c|}{P} & \multicolumn{1}{c|}{SS} & \multicolumn{1}{c|}{83.1 (3.8)}         & \multicolumn{1}{c|}{89.9 (1.6)}         & \multicolumn{1}{c|}{71.4 (3.5)}         & \multicolumn{1}{c|}{87.8 (1.3)}                  & \multicolumn{1}{c|}{94.3 (0.8)}                  & \multicolumn{1}{c}{65.4 (2.6)}                  \\
\multicolumn{1}{c|}{IH} & \multicolumn{1}{c|}{P} & \multicolumn{1}{c|}{AR} & \multicolumn{1}{c|}{83.1 (3.7)}         & \multicolumn{1}{c|}{91.6 (2.7)}         & \multicolumn{1}{c|}{92.9 (1.3)}         & \multicolumn{1}{c|}{\textbf{92.9 (0.8)}}         & \multicolumn{1}{c|}{92.7 (0.7)}                  & \multicolumn{1}{c}{80.3 (2.5)}                  \\
\multicolumn{1}{c|}{IH} & \multicolumn{1}{c|}{P} & \multicolumn{1}{c|}{UU} & \multicolumn{1}{c|}{83.1 (3.8)}         & \multicolumn{1}{c|}{91.5 (3.3)}         & \multicolumn{1}{c|}{95.5 (0.7)}         & \multicolumn{1}{c|}{\textbf{97.0 (0.5)}}         & \multicolumn{1}{c|}{80.1 (2.5)}                  & \multicolumn{1}{c}{93.5 (0.6)}                  \\
\multicolumn{1}{c|}{IH} & \multicolumn{1}{c|}{P} & \multicolumn{1}{c|}{UC} & \multicolumn{1}{c|}{83.1 (3.8)}         & \multicolumn{1}{c|}{95.8 (0.8)}         & \multicolumn{1}{c|}{78.8 (3.8)}         & \multicolumn{1}{c|}{95.7 (1.8)}                  & \multicolumn{1}{c|}{96.2 (1.0)}                  & \multicolumn{1}{c}{\textbf{97.7 (0.3)}}         \\
\bottomrule
\end{tabular}
}
\caption{\footnotesize{Evaluation results on English tweets reported across active learning methods -- stratified sampling ('SS'), adaptive retrieval ('AR'), uncertainty uncalibrated ('UU'), uncertainty calibrated ('UC'), exploit-explore retrieval ('EE') -- performance metrics -- average precision ('P'), number of predicted positives ('E'), diversity ('D') -- and classes -- is hired ('IH'), is unemployed ('IU'), job offer ('JO'), job search ('JS'), job loss ('LJ'). Standard errors for P and D are shown in parentheses, and we report a lower bound and an upper bound for E. Bold values indicate the iteration at which a model converges.}}
\label{tab:avg_precision_cross_method}
\end{table*}